\documentclass{article}

\usepackage[final]{neurips_2023}

\usepackage[utf8]{inputenc}
\usepackage[T1]{fontenc}
\usepackage{hyperref}
\usepackage{url}
\usepackage{booktabs}
\usepackage{amsfonts}
\usepackage{amsmath}
\usepackage{nicefrac}
\usepackage{microtype}
\usepackage{xcolor}
\usepackage{graphicx}
\usepackage{tabularx}
\usepackage{array}
\usepackage{multirow}
\usepackage{makecell}
\usepackage{fvextra}
\usepackage{float}
\usepackage{placeins}
\usepackage{seqsplit}
\usepackage{etoolbox}

\hypersetup{
  colorlinks=false,
  linkbordercolor={0 1 0},
  citebordercolor={0 1 0},
  urlbordercolor={0 1 0},
  pdfauthor={Zihan Qin, Boao Xu, Zhao Dong, Yingping Sun, Ziheng Jiao, Junying Wang, Hongwei Wang},
  pdftitle={RS-RIE-Bench: Benchmarking Reasoning-Guided Remote Sensing Image Editing}
}

\makeatletter
\AtBeginEnvironment{thebibliography}{\sloppy}
\renewcommand{\@notice}{}
\makeatother

\title{RS-RIE-Bench: Benchmarking Reasoning-Guided Remote Sensing Image Editing}

\author{%
  Zihan Qin$^{1,\dagger}$\\
  \texttt{qzh2025@mai1.nwpu.edu.cn}
  \And
  Boao Xu$^{1,\dagger}$\\
  \texttt{xu.paper@mai1.nwpu.edu.cn}
  \And
  Zhao Dong$^{2}$\\
  \texttt{casper\_dong@126.com}
  \AND
  Yingping Sun$^{3}$\\
  \texttt{sunyingping510@163.com}
  \And
  Ziheng Jiao$^{4}$\\
  \texttt{jiaoziheng1@huawei.com}
  \AND
  Junying Wang$^{5,*}$\\
  \texttt{junyingwang25@m.fudan.edu.cn}
  \And
  Hongwei Wang$^{1}$\\
  \texttt{wanghongwei@nwpu.edu.cn}
}

\begin{document}

\maketitle

\begin{abstract}
Remote sensing image editing aims to modify remote sensing images according to natural language instructions while preserving geographic rules and sensor observation characteristics. Existing benchmarks mainly target natural images or general visual scenes, and thus may not fully capture the reasoning, regional control, and sensor-consistency abilities required in remote sensing editing. To fill this gap, we introduce RS-RIE-Bench, the first benchmark for reasoning-guided remote sensing image editing. RS-RIE-Bench organizes tasks into three categories: temporal reasoning, causal reasoning, and spatial reasoning. These categories capture temporal evolution, causal consequence, and spatial imaging consistency in remote sensing scenes. The evaluation protocol covers three dimensions: target region plausibility, non-target region preservation, and image quality consistency. We further demonstrate the feasibility of MLLM-based evaluation through cross-judge consistency analysis and stratified expert review. Systematic evaluation on eight open-source and closed-source image editing models shows that current models still have clear limitations in reasoning-guided remote sensing editing. Even the strongest model achieves only 24.28\% overall accuracy under the strict joint-satisfaction criterion, while the mean relaxed joint-4 success rate across all eight models is 32.23\%. Causal reasoning and spatial reasoning remain especially challenging, and several open-source models are close to zero in some categories. These results show that RS-RIE-Bench can effectively reveal the limitations of current models in geographic reasoning, regional control, and sensor-consistent generation. It also provides a standardized benchmark and a clear research direction for future remote sensing intelligent editing models.
\end{abstract}

\begingroup
\renewcommand{\thefootnote}{\fnsymbol{footnote}}
\footnotetext[1]{Corresponding author.}
\footnotetext[2]{Equal contribution.}
\endgroup
\begingroup
\renewcommand{\thefootnote}{\arabic{footnote}}
\footnotetext[1]{School of Artificial Intelligence, Optics and Electronics (iOPEN), Northwestern Polytechnical University, Xi'an 710072, China.}
\footnotetext[2]{Xi'an Modern Control Technology Research Institute, Xi'an 710065, China.}
\footnotetext[3]{Lanzhou Institute of Physics, Lanzhou 730000, China.}
\footnotetext[4]{Huawei Technologies Co., Ltd., China.}
\footnotetext[5]{Fudan University and Shanghai Artificial Intelligence Laboratory, Shanghai 200000, China.}
\endgroup

\section{Introduction}

Remote sensing image editing aims to modify Earth observation images according to natural language instructions while preserving the geographic layout and sensor-specific appearance of the original scene~\cite{brooks2023instructpix2pix,wang2023editbench}. Unlike natural-image editing, where visual plausibility is often sufficient, remote sensing editing is constrained by geospatial topology, scale-dependent object appearance, illumination geometry, and sensor-specific texture statistics~\cite{kuckreja2024geochat,luo2026vlrs,chen2026rsedit}. A visually plausible edit may still be unusable if it shifts roads, changes background land cover, violates shadow geometry, or replaces the original sensor style with a synthetic rendering. These constraints make remote sensing image editing a joint test of semantic instruction following, geospatial reasoning, regional control, and sensor-consistent generation~\cite{kuckreja2024geochat,chen2026rsedit}. This task therefore has important value in disaster response, environmental monitoring, urban planning, and land surface change analysis.

Recent advances in large multimodal models have also driven the evolution of image editing benchmarks~\cite{li2023blip2,dai2023instructblip}. Early benchmarks, such as EditBench~\cite{wang2023editbench} and MagicBrush~\cite{zhang2023magicbrush}, mainly focused on explicit editing instructions, such as object removal or local attribute replacement. As models became stronger, benchmarks for reasoning-driven image editing started to emerge. RISE-Bench~\cite{zhao2026risebench} was the first benchmark to systematically evaluate reasoning ability in image editing, marking an important step toward reasoning-driven editing evaluation. GRADE~\cite{liu2026grade} extended evaluation from common visual scenes to domain knowledge settings and showed that domain knowledge makes the task much harder. KrisBench~\cite{wu2026krisbench} organized editing tasks from the perspective of cognitive taxonomy and provided a structured framework for reasoning-driven editing. These studies suggest that reasoning ability has become an important dimension for distinguishing image editing models, beyond low-level generation quality alone.

\begin{figure}[!t]
    \centering
    \includegraphics[width=0.92\textwidth]{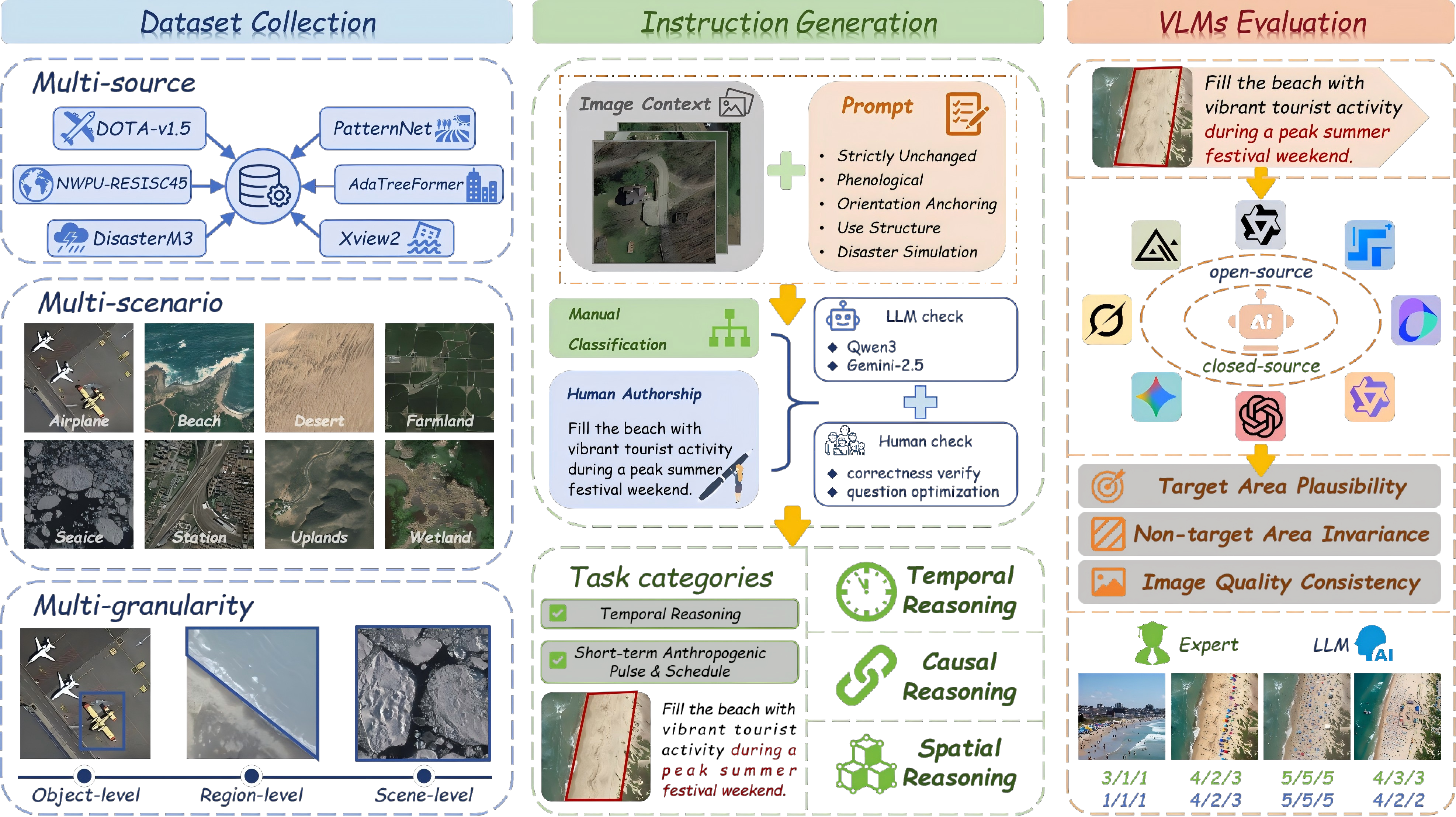}
    \caption{Overview of RS-RIE-Bench. The benchmark contains 486 samples across three task categories and evaluates model outputs with target region plausibility, non-target region preservation, and image quality consistency.}
    \label{fig:framework}
\end{figure}

However, existing reasoning-guided image editing benchmarks mainly target natural images or general academic scenarios~\cite{zhao2026risebench,liu2026grade,wu2026krisbench}. The missing evaluation setting is not simply remote sensing imagery alone, but the combination of reasoning-guided editing, geospatial constraint satisfaction, non-target preservation, and sensor-style consistency~\cite{kuckreja2024geochat,luo2026vlrs,chen2026rsedit,wang2026disasterm3}. A benchmark for this setting must test whether a model can infer the intended land-surface or imaging transformation, apply it to the correct region, preserve irrelevant geographic structures, and maintain the visual statistics of the original sensor observation. As illustrated by the failure example in Figure~\ref{fig:framework}, an output may appear semantically plausible at first glance, yet still fail because it disturbs background topology or replaces the native remote sensing texture with a re-rendered style. To fill this gap, we introduce RS-RIE-Bench as a benchmark for reasoning-guided remote sensing image editing. RS-RIE-Bench divides editing tasks into three reasoning categories: temporal reasoning, causal reasoning, and spatial reasoning. These categories capture process-driven evolution, event-triggered change, and spatial presentation under sensor imaging conditions.

At the evaluation level, RS-RIE-Bench further builds a three-dimensional evaluation protocol that matches the needs of remote sensing editing. First, target region plausibility evaluates whether the edited region follows the temporal evolution, causal consequence, or spatial constraint implied by the instruction. In other words, it tests whether the model can generate a geographically plausible target change. Second, non-target region preservation evaluates whether regions that are not supposed to be edited remain stable in both semantics and appearance. This is especially important in continuous land-cover scenes with roads, buildings, water bodies, and vegetation, where precise regional control is difficult and unauthorized changes can easily appear. Third, image quality consistency evaluates whether the edited result preserves the sensor-level realism and imaging consistency of remote sensing images. This includes texture continuity, radiometric consistency, geometric coordination, and the absence of obvious generation artifacts, over-enhancement, or stylized traces. To support scalable and reliable evaluation, we combine MLLM-as-a-Judge with expert review and verify the reliability of the automatic protocol through consistency analysis.

In the experiments, we systematically test eight open-source and closed-source image editing models. The results show that current models still have clear limitations in reasoning-guided remote sensing editing. Even the strongest model achieves only 24.28\% overall accuracy under the strict joint-satisfaction criterion. Causal reasoning and spatial reasoning remain especially difficult, and several open-source models are close to zero in some categories. These findings show that RS-RIE-Bench can effectively reveal the limitations of current models in geographic reasoning, spatial control, and sensor-consistent generation. It provides a standardized reference for the design and evaluation of future remote sensing intelligent editing models.

The main contributions of this paper are as follows:
\begin{enumerate}
    \item We define a geospatial reasoning taxonomy for remote sensing image editing, covering temporal land-surface evolution, event-driven causal change, and spatial imaging transformation.
    \item We construct RS-RIE-Bench, a benchmark of 486 reasoning-guided remote sensing image editing tasks with explicit task metadata, source-dataset provenance, instruction annotations, and a quality-control workflow.
    \item We propose a three-dimensional evaluation protocol that separately measures target region plausibility, non-target region preservation, and image quality consistency, and we validate this protocol using cross-judge consistency and stratified expert review.
    \item We benchmark eight current image editing systems and identify systematic failure modes in semantic edit execution, geospatial preservation, and sensor-consistent generation.
\end{enumerate}

\section{Related Works}

\subsection{Image Generation and Editing Models}

Early image synthesis frameworks mainly relied on latent diffusion models, represented by the Stable Diffusion series~\cite{rombach2022ldm}. Their core idea is to guide the denoising process in latent space with text prompts. In recent years, this paradigm has gradually shifted toward flow matching and Transformer-based architectures~\cite{esser2024sd3,labs2025flux}. Stable Diffusion 3 systematically explored the scaling behavior of rectified flow Transformers for high-resolution image synthesis~\cite{esser2024sd3}. FLUX further demonstrated the practical potential of large-scale flow matching for in-context image generation and editing in latent space~\cite{labs2025flux}. These advances significantly improved image fidelity and prompt alignment. At the same time, the field has moved toward unified multimodal models~\cite{openai2025gpt5,comanici2025gemini25,deng2025bagel}. New native multimodal large models, such as GPT-5~\cite{openai2025gpt5}, Gemini 2.5 Flash~\cite{comanici2025gemini25}, and BAGEL~\cite{deng2025bagel}, no longer separate understanding from generation. Instead, they use a unified token stream across modalities for joint training~\cite{deng2025bagel}. This end-to-end design allows the model to form a closed loop between complex visual context understanding and multimodal image generation within a shared representation space.

Alongside this progress in generation, image editing has evolved from early mask-based local inpainting to flexible instruction-following through natural language. InstructPix2Pix~\cite{brooks2023instructpix2pix} pioneered this direction and showed that images can be edited directly from language instructions without explicit masks. More recent work has moved further toward deeper integration between multimodal large language model architectures and diffusion Transformers~\cite{xiao2025omnigen,wu2025qwenimage,liu2025step1x,deng2025bagel}. Models such as OmniGen~\cite{xiao2025omnigen}, Qwen-Image~\cite{wu2025qwenimage}, Step1X-Edit~\cite{liu2025step1x}, and BAGEL~\cite{deng2025bagel} incorporate native reasoning-before-editing pipelines. This design allows them to use continuous vision-language context for complex free-form and multi-image editing. Despite their strong multimodal composition ability, current editing models are almost entirely trained and optimized on natural image datasets~\cite{brooks2023instructpix2pix,zhang2023magicbrush}. Their ability to understand and manipulate remote sensing images remains largely unexplored. This leaves a major technical gap when open-world instruction editing is applied to overhead geospatial scenes.

\subsection{Image Editing Benchmarks}

Early benchmarks, such as EditBench~\cite{wang2023editbench} and MagicBrush~\cite{zhang2023magicbrush}, were mainly designed to verify explicit semantic instructions, such as color modification or object replacement. These tasks rely more on low-level perception than on deep cognitive reasoning. To evaluate higher-level intelligence, recent studies have started to build reasoning-centered benchmarks. RISE-Bench~\cite{zhao2026risebench} introduced the first reasoning-driven image editing benchmark. It organized 360 carefully selected samples into four cognitive dimensions: temporal, causal, spatial, and logical reasoning. Along the same direction, KrisBench~\cite{wu2026krisbench} built its evaluation framework from a cognitive taxonomy and tested how well models align with human common sense in open-world transformations. Moving beyond daily common sense toward more rigorous expert settings, GRADE~\cite{liu2026grade} proposed a discipline-informed benchmark with 520 samples across ten academic domains. It adopted a multi-dimensional evaluation protocol to assess domain reasoning, visual consistency, and logical readability.

Although these frameworks successfully benchmark implicit reasoning ability, they are still limited to natural images, daily common sense, or abstract academic figures. There is still no benchmark for reasoning-guided editing in the remote sensing domain. In remote sensing image editing, the task is uniquely constrained by strict Earth geometry, atmospheric conditions, and complex geophysical or sensor-physics constraints. This gap motivates the need for a benchmark dedicated to reasoning-guided remote sensing image editing.

To clarify the scope of RS-RIE-Bench, Table~\ref{tab:benchmark-comparison} compares representative benchmarks along the dimensions most relevant to reasoning-guided remote sensing editing. Existing benchmarks cover either general-domain image editing, natural-image reasoning, or remote-sensing understanding tasks. However, they do not jointly evaluate geospatially constrained instruction editing, non-target region preservation, and sensor-style consistency in remote sensing imagery.

\begin{table}[!t]
    \centering
    \small
    \caption{Comparison of representative benchmarks related to reasoning-guided remote sensing image editing.}
    \label{tab:benchmark-comparison}
    \resizebox{\textwidth}{!}{%
    \begin{tabular}{lcccccc}
        \toprule
        Benchmark & Domain & Editing & Reasoning & Remote-sensing Constraints & Non-target Preservation & Sensor Consistency \\
        \midrule
        EditBench~\cite{wang2023editbench} & Natural images & Yes & Limited & No & Limited & No \\
        MagicBrush~\cite{zhang2023magicbrush} & Natural images & Yes & Limited & No & Limited & No \\
        RISE-Bench~\cite{zhao2026risebench} & Natural images & Yes & Yes & No & Partial & No \\
        GRADE~\cite{liu2026grade} & Academic/domain figures & Yes & Yes & No & Partial & No \\
        GeoChat~\cite{kuckreja2024geochat} / VLRS-Bench~\cite{luo2026vlrs} & Remote sensing & No & Yes & Yes & N/A & N/A \\
        RS-RIE-Bench & Remote sensing & Yes & Yes & Yes & Yes & Yes \\
        \bottomrule
    \end{tabular}}
\end{table}

\subsection{Remote Sensing Image Analysis and MLLM-as-a-Judge}

The remote sensing community has already built relatively mature benchmarks and model foundations for vision-language understanding and interpretation tasks, such as land-cover classification, object detection, and change detection. Representative examples include GeoChat~\cite{kuckreja2024geochat}, VLRS-Bench~\cite{luo2026vlrs}, and DisasterM3~\cite{wang2026disasterm3}. In parallel, generative work in remote sensing has begun to explore text-guided editing and change simulation. RSEdit~\cite{chen2026rsedit} studies domain-specialized text-guided remote sensing image editing, while Changen~\cite{zheng2023changen} focuses on scalable multi-temporal change data generation for remote sensing analysis. These efforts are important for model construction and data augmentation, but they do not provide a benchmark that jointly evaluates instruction-grounded editing, non-target preservation, and sensor-style consistency under remote sensing constraints. There is still no systematic framework that can execute and evaluate reasoning-guided editing, for example, testing whether a model can simulate and edit geopolitically or ecologically driven scene changes in Earth observation imagery from an abstract instruction.

MLLM-as-a-Judge has become a practical tool for scalable evaluation of open-ended visual generation~\cite{zheng2023judging} and has been adopted in automatic pipelines such as RISE-Bench~\cite{zhao2026risebench} and GRADE~\cite{liu2026grade}, but it should not be treated as ground truth. Its reliability depends on prompt design, domain alignment, calibration against expert judgment, and robustness across judge models. This issue is particularly important in remote sensing, where evaluation often requires geospatial and sensor-specific knowledge that may not be captured by a general-purpose judge. Therefore, directly using a general open-world judge model in the remote sensing domain leads to serious domain mismatch, and customized judge prompt design that integrates Earth science knowledge remains essential for accurate and scalable evaluation of reasoning-guided remote sensing image editing.

\section{Framework}

This section introduces the overall framework of RS-RIE-Bench. We first describe its hierarchical task design, which includes three core reasoning categories and nine fine-grained dimensions. We then present the data collection process and the quality control criteria to ensure reliability, representativeness, and challenge. Finally, we describe the three-dimensional evaluation pipeline that supports a comprehensive assessment of image editing quality.

\begin{figure}[!t]
    \centering
    \includegraphics[width=0.95\textwidth]{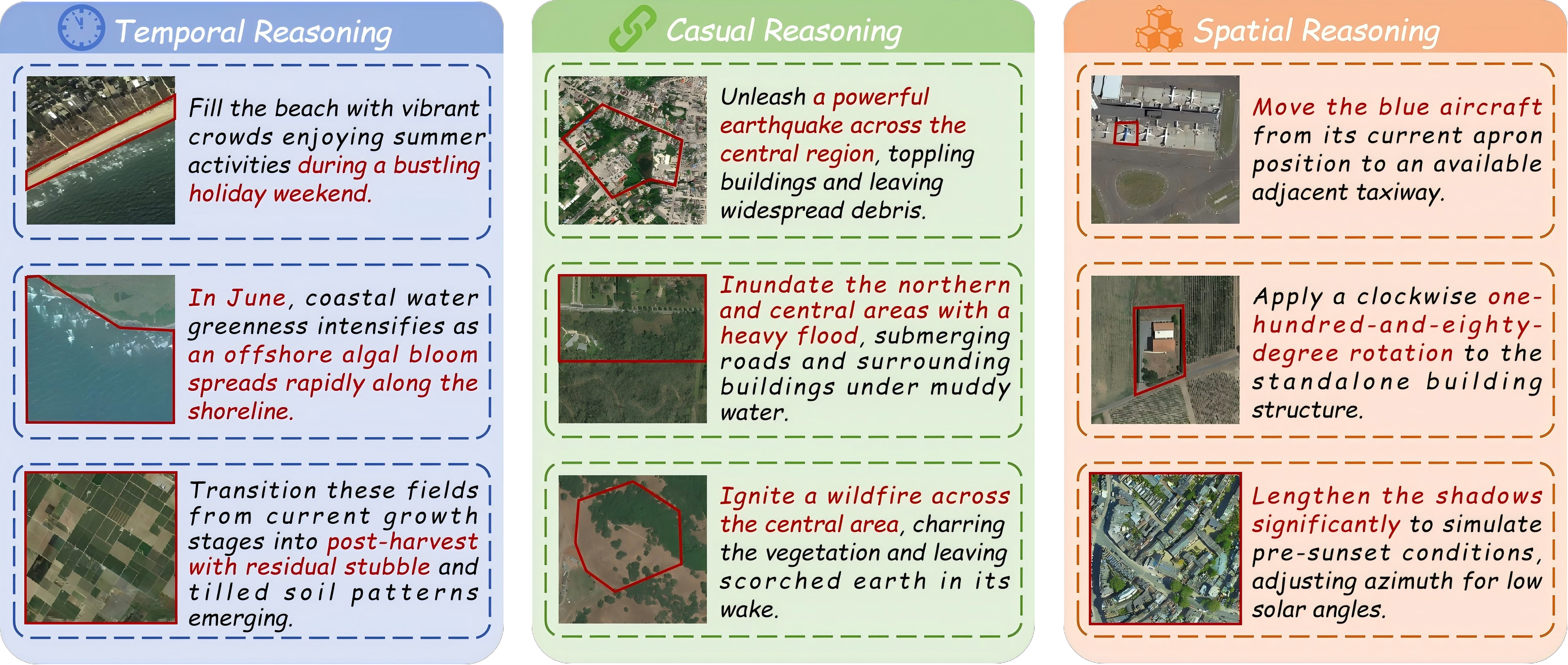}
    \caption{Representative reasoning-guided remote sensing image editing cases in RS-RIE-Bench. The examples are drawn from the benchmark's three categories and nine fine-grained dimensions, and each case requires the model to infer how the scene should change rather than follow an explicit editing operation.}
    \label{fig:reasoning-case}
\end{figure}

\subsection{Reasoning Taxonomy}
\label{sec:reasoning-taxonomy}

\begin{figure}[!t]
    \centering
    \includegraphics[width=0.4\textwidth]{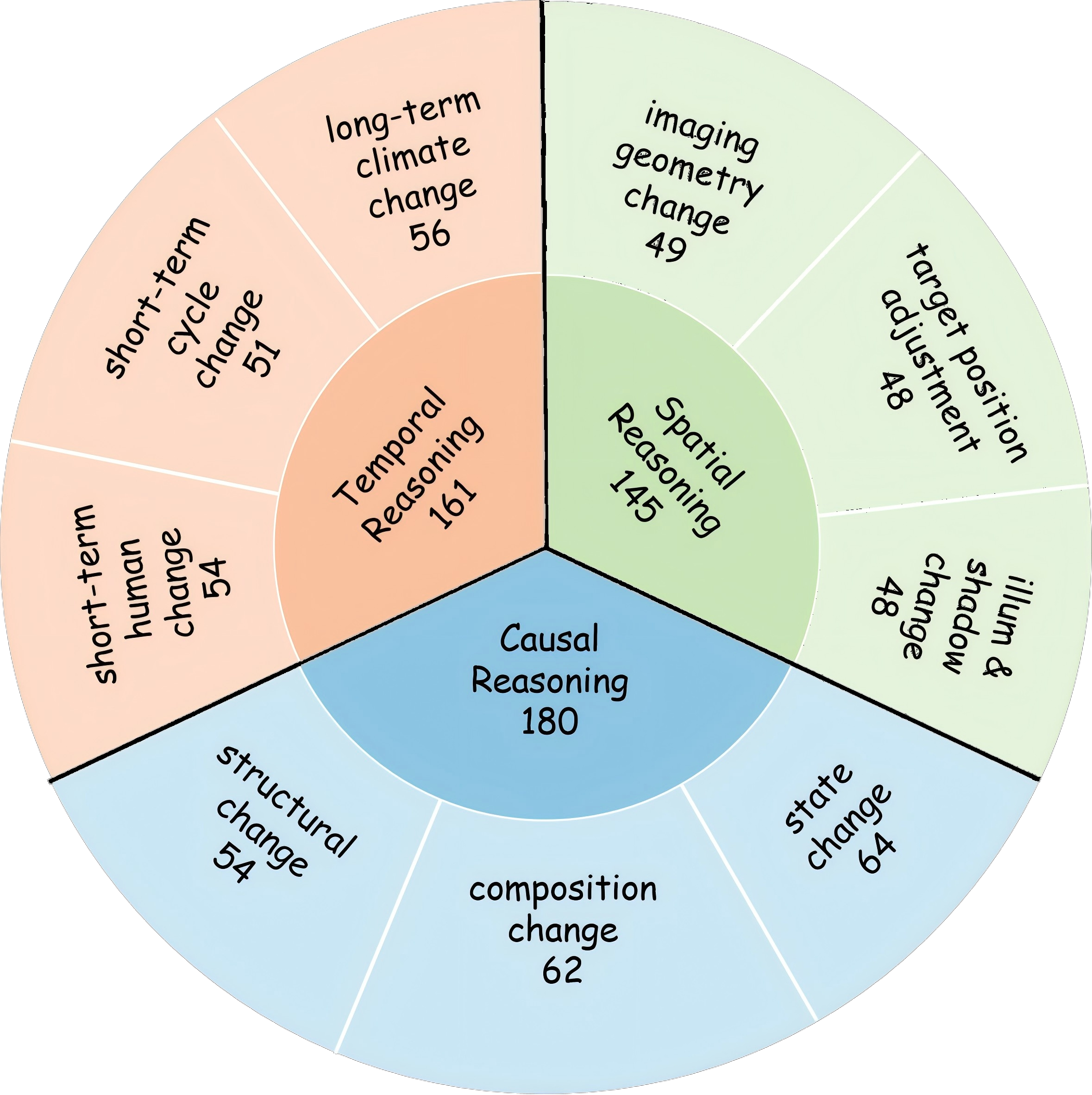}
    \caption{Distribution of RS-RIE-Bench. The benchmark contains 486 samples in total, including 161 temporal reasoning samples (33.13\%), 180 causal reasoning samples (37.04\%), and 145 spatial reasoning samples (29.84\%).}
    \label{fig:category}
\end{figure}

Among the many remote sensing image editing tasks, RS-RIE-Bench focuses on three major task categories that require deep visual understanding and precise reasoning: temporal reasoning, causal reasoning, and spatial reasoning. These categories capture temporal evolution, causal consequence, and spatial imaging consistency. For each category, we carefully design and collect diverse high-quality test cases. Each case contains an input image and an editing instruction, and illustrates a reasoning-based image transformation process, as shown in Figure~\ref{fig:reasoning-case}. Figure~\ref{fig:category} summarizes the benchmark distribution across the three categories and nine fine-grained dimensions.

Temporal reasoning mainly evaluates whether the model can generate reasonable edited results by following temporal evolution patterns on the land surface. Beyond recognizing static attributes such as color, shape, or size, a reasoning-capable generative model should understand how land objects or scenes change over time. In remote sensing scenarios, such changes are not arbitrary. They form a continuous process constrained by three key factors: the direction of change, the rate of evolution, and the scale of influence. Based on this observation, we divide temporal reasoning into three subcategories: short-term natural cycle change, short-term human activity change, and long-term climate-driven change. Together, these subcategories cover diverse temporal change scenarios under natural processes, human activities, and long-term climate effects. This design supports a more systematic evaluation of temporal reasoning under different change mechanisms and evolution processes.

Causal reasoning evaluates how well the model understands and expresses causal chains. The focus is on how a given event or condition is converted into concrete visual changes on the land surface. Unlike temporal reasoning, which emphasizes continuous evolution, causal reasoning emphasizes how external forces or events trigger land surface state changes and then produce observable consequences. We further divide causal reasoning into three subcategories: structural change, state change, and composition change. Specifically, structural change focuses on the damage of object form and spatial structure. State change focuses on changes in cover type or physical condition. Composition change focuses on deeper changes in material composition and biochemical properties. These three subcategories progress from form to state to composition and together cover common types of causal change in remote sensing imagery.

Spatial reasoning mainly concerns how well the model understands and expresses spatial relations in remote sensing imaging. In remote sensing images, whether the change happens in a local area or extends to a larger region, the edited result must stay consistent with the projection relation, illumination direction, and spatial topology of the whole image. This spatial consistency is reflected in three aspects: imaging geometry change, illumination and shadow change, and target position and spatial relation adjustment. Specifically, imaging geometry change focuses on projection deformation and occlusion changes caused by sensor viewpoint variation. Illumination and shadow change focuses on joint changes in shadow direction, length, and brightness caused by different solar illumination conditions. Target position and spatial relation adjustment focuses on whether spatial relations between the edited target and the surrounding scene remain coordinated after target movement, insertion, or removal. Together, these aspects show that spatial reasoning is not only about moving a target itself. More importantly, it is about maintaining coherence between the edited result, the global imaging condition, and the spatial organization of the scene.

\subsection{Data Collection and Quality Control}

The construction of RS-RIE-Bench follows a five-step filtering funnel: candidate collection, quality screening, taxonomy assignment, human review, and final retention. We begin with 521 pre-screened RGB optical remote sensing candidates from the raw image pool, including 166 temporal, 205 causal, and 150 spatial images. During taxonomy assignment, 35 ambiguous or multi-label cases are removed, leaving 486 final samples: 161 temporal, 180 causal, and 145 spatial. These stages constrain the source of the samples, the content of the tasks, and the final retention criteria, ensuring the representativeness and reliability of the benchmark. We collect candidate images from six public RGB optical remote sensing sources, namely DOTA-v1.5~\cite{xia2018dota}, PatternNet~\cite{zhou2018patternnet}, NWPU-RESISC45~\cite{cheng2017nwpu}, AdaTreeFormer-related tree-counting datasets~\cite{amirkolaee2024adatreeformer}, DisasterM3~\cite{wang2026disasterm3}, and xView2/xBD~\cite{gupta2019xbd}.\footnote{DOTA-v1.5 is distributed via the official DOTA page (\url{https://captain-whu.github.io/DOTA/dataset.html}); the page specifies Google Earth, GF-2/JL-1, and CycloMedia as image sources but does not state a standalone dataset license. PatternNet is publicly released at \url{https://sites.google.com/view/zhouwx/dataset}; its images are collected from Google Earth or the Google Map API, and the official homepage does not state a unified redistribution license. NWPU-RESISC45 is publicly available at \url{http://www.escience.cn/people/JunweiHan/NWPU-RESISC45.html}; the dataset is also mirrored on Figshare under CC BY 4.0, and the images are RGB scenes extracted from Google Earth. AdaTreeFormer uses the official repository at \url{https://github.com/HAAClassic/AdaTreeFormer}, which provides download links for the underlying Jiangsu, Yosemite, and London tree-counting datasets, but no unified dataset-wide license is stated. DisasterM3 is released through the official repository at \url{https://github.com/Junjue-Wang/DisasterM3}; the repository states that the images and annotations are for academic use only. xView2/xBD is described on the official challenge page \url{https://www.sei.cmu.edu/projects/xview-2-challenge/} and the dataset page \url{https://xview2.org/dataset}; the SEI states that xBD is available for public use under a Creative Commons license, and the baseline repository reports CC BY-NC-SA 4.0.} These sources cover object-level, scene-level, and disaster-oriented imagery, providing broad diversity in land surface environments and spatial scales.

Each manually written instruction is reviewed by \texttt{gpt-4o}~\cite{openai2024gpt4o} and \texttt{gemini-2.5}~\cite{comanici2025gemini25} for ambiguity and category consistency, and disagreements are resolved by expert adjudication from annotators with remote-sensing or image-interpretation backgrounds. The matching instruction is manually written to express the required event, condition, or state change rather than an explicit low-level editing operation. The mean instruction lengths are 30.19 words for temporal reasoning, 30.43 words for causal reasoning, and 29.41 words for spatial reasoning, with an overall range of 22--40 words. This keeps the task focused on a clear reasoning objective while maintaining comparable linguistic complexity across categories.

During quality control, we mainly ensure realism, certainty, and evaluability. First, all source images should have clear and interpretable imaging quality without obvious artifacts. The target change should also appear at a spatial scale suitable for editing and evaluation. Second, all editing instructions should contain sufficiently clear geoscience or imaging constraints so that the interpretation space does not become too large. Through this process, we try to reduce data noise, task ambiguity, and evaluation drift as much as possible, thereby improving the stability and reliability of the benchmark.

\subsection{Evaluation Pipeline}
\label{sec:evaluation-pipeline}

The effectiveness of remote sensing image editing cannot be fully captured by a single metric. An edited result may appear to follow the instruction, yet still violate geoscientific rules in the target region, introduce unnecessary disturbance in non-target regions, or produce sensor-inconsistent artifacts in the full image. Therefore, RS-RIE-Bench adopts a three-dimensional evaluation pipeline. It jointly measures model outputs from target region plausibility, non-target region preservation, and image quality consistency, so that the real capability of a model on remote sensing editing tasks can be assessed more comprehensively.

Target region plausibility evaluates whether the model generates changes in the edited region that follow geoscientific rules. In remote sensing image editing, changes in the target region should not merely look semantically reasonable. They should also be interpretable in terms of land surface processes, physical structures, or environmental mechanisms. For example, building appearance after a disaster, vegetation state under seasonal variation, and land surface reconstruction caused by long-term environmental evolution should all follow domain-specific rules. Based on this principle, this dimension mainly evaluates whether the edited result expresses a correct and plausible change in the target region.

Non-target region preservation evaluates whether the model can avoid unnecessary disturbance to irrelevant regions while following the editing instruction. Compared with explicit object boundaries in natural images, land objects in remote sensing images are often continuously distributed and have vague boundaries. Clear separation between target and non-target regions is often missing. As a result, local editing is more likely to introduce unintended changes outside the target area. This dimension measures whether the model has fine-grained regional control, that is, whether it can apply changes only where necessary while keeping other regions stable in both content and appearance.

Image quality consistency evaluates whether the edited result preserves the sensor imaging characteristics expected in remote sensing imagery. Unlike general image generation tasks, remote sensing images have specific resolution characteristics, texture statistics, and imaging styles. Therefore, high-quality editing does not mean a more attractive or sharper image. Instead, it means that the result does not introduce generation artifacts, unauthorized AI enhancement, or visual expressions that are inconsistent with the original observation condition. This dimension constrains the model to preserve the realism and usability of remote sensing images while completing the edit.

\begin{figure}[!t]
    \centering
    \includegraphics[width=0.98\textwidth]{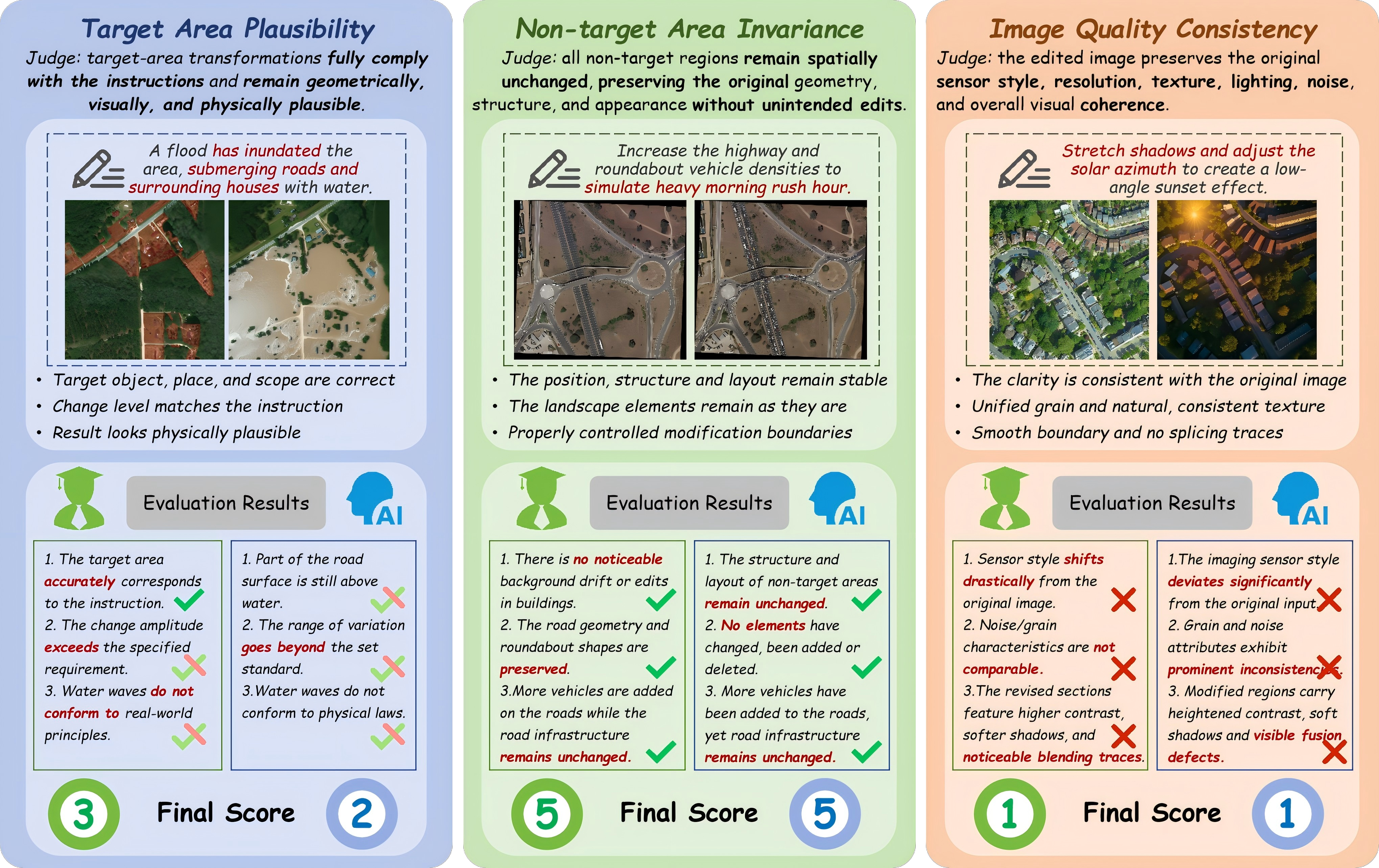}
    \caption{Representative examples of the three evaluation dimensions in RS-RIE-Bench. The figure illustrates how target region plausibility, non-target region preservation, and image quality consistency are assessed in practice.}
    \label{fig:evaluation-case}
\end{figure}

Figure~\ref{fig:evaluation-case} illustrates these three dimensions with representative examples. The target-region panel highlights whether the edited area follows the instruction and remains geoscientifically plausible, the non-target-region panel checks whether irrelevant areas stay unchanged, and the image-quality panel shows whether the output preserves the original sensor style and visual coherence.

In the actual evaluation process, we use MLLM-as-a-Judge as a scalable proxy evaluator rather than ground truth. The primary proxy judge, \texttt{gpt-5.1}~\cite{openai2024gpt51}, scores model outputs on the three dimensions above under a fixed rubric, and we use expert review as an external validation and calibration signal for the proxy. For summary metrics, RS-RIE-Bench adopts strict joint-satisfaction accuracy as the core criterion. A sample is counted as correct only when it satisfies all three evaluation dimensions at the same time. This design provides a consistent and scalable way to compare models while still acknowledging that automatic judge scores are an approximation of expert judgment rather than the final truth.

\section{Experiments}

We systematically evaluate the performance of representative image editing models on RS-RIE-Bench. This section analyzes the capability boundary of current models in reasoning-guided remote sensing image editing from three aspects: main results, evaluation consistency, and the relations among evaluation dimensions.

\subsection{Experimental Setup}

We evaluate eight representative image editing models using fixed model versions and a unified generation protocol. The closed-source models are \texttt{gpt-image-2}~\cite{openai2026gptimage2}, \texttt{\seqsplit{gemini-3.1-flash-image-preview}}~\cite{google2026gemini31flashimage}, \texttt{\seqsplit{wan2.7-image-pro}}~\cite{mao2026wanimage}, \texttt{grok-imagine-image}~\cite{xai2026grokimagineimage}, and \texttt{\seqsplit{doubao-seedream-5-0-260128}}~\cite{gao2025seedream3}. The open-source models are \texttt{Flux.2-dev}~\cite{labs2025flux}, \texttt{Step1X-Edit}~\cite{liu2025step1x}, and \texttt{Qwen-Image-Edit-2509}~\cite{wu2025qwenimage}, and they are run with the released checkpoints under the same input images, prompt templates, output-resolution settings, and post-processing rules whenever supported. These evaluated models cover both unified multimodal models and specialized image editing models. We select this set as a practical and reproducible baseline suite for reasoning-guided remote sensing image editing. The selected models cover widely used closed-source systems and publicly available open-source editors, and they span a broad range of capability levels. This makes it possible to observe how RS-RIE-Bench distinguishes representative baselines across different capability boundaries under a unified protocol. By fixing the model versions and applying the same input, output, and post-processing settings whenever supported, we ensure that the comparison remains fair and reproducible.

The main evaluation metric follows the strict joint-satisfaction protocol defined in Section~\ref{sec:evaluation-pipeline}. For automatic evaluation, we uniformly use \texttt{gpt-5.1}~\cite{openai2024gpt51} as the primary proxy judge under a fixed scoring prompt and a three-dimensional rubric. Specifically, the proxy judge assigns each edited result a score from 1 to 5 on target region plausibility, non-target region preservation, and image quality consistency, with explicit criteria for each level provided in the appendix. A sample is counted as successful only when all three dimensions receive a score of 5.

Beyond the main results, we further examine the stability and validity of the proxy-based evaluation protocol. Specifically, we compare the scoring consistency of \texttt{gpt-5.1}~\cite{openai2024gpt51}, \texttt{gemini-3-flash-preview}~\cite{google2026gemini3flashpreview}, and \texttt{qwen-3.5}~\cite{qwen2026qwen35omni} on the same anonymized set of model outputs using identical input formatting and the same rubric. We also compare the scores from \texttt{gpt-5.1} with expert review results, where human review serves as an external validation rather than the primary metric. In addition, we compute pairwise correlations among the three evaluation dimensions on the full set of proxy scores to examine whether the dimensions provide complementary information.

\begin{table}[!t]
    \centering
    \small
    \caption{Main results on RS-RIE-Bench. The category columns report strict joint-satisfaction accuracy (\%), while the two overall columns report the relaxed joint-4 accuracy as mean $\pm$ approximate Wilson 95\% half-width~\cite{wilson1927probable}. A sample is counted as successful only when target region plausibility, non-target region preservation, and image quality consistency all receive a score of 5 for joint-5, or at least 4 for joint-4.}
    \label{tab:main-results}
    \resizebox{\textwidth}{!}{%
    \begin{tabular}{lcccccc}
        \toprule
        Model & Type & Temporal & Causal & Spatial & Overall (J-5) & Overall (J-4) \\
        \midrule
        \texttt{gpt-image-2}~\cite{openai2026gptimage2} & Closed & 21.08 & 17.56 & 32.00 & 24.28 $\pm$ 3.80 & 61.32 $\pm$ 4.22 \\
        \texttt{gemini-3.1-flash-image-preview}~\cite{google2026gemini31flashimage} & Closed & 6.63 & 17.07 & 10.67 & 12.76 $\pm$ 2.97 & 44.03 $\pm$ 4.40 \\
        \texttt{wan2.7-image-pro}~\cite{mao2026wanimage} & Closed & 3.01 & 3.90 & 11.33 & 5.97 $\pm$ 2.13 & 37.86 $\pm$ 4.30 \\
        \texttt{grok-imagine-image}~\cite{xai2026grokimagineimage} & Closed & 4.82 & 0.98 & 3.33 & 3.09 $\pm$ 1.58 & 14.61 $\pm$ 3.14 \\
        \texttt{doubao-seedream-5-0-260128}~\cite{gao2025seedream3} & Closed & 19.02 & 22.77 & 19.46 & 20.37 $\pm$ 3.58 & 45.88 $\pm$ 4.45 \\
        \texttt{Flux.2-dev}~\cite{labs2025flux} & Open & 6.63 & 1.95 & 12.67 & 6.79 $\pm$ 2.25 & 33.95 $\pm$ 4.19 \\
        \texttt{Step1X-Edit}~\cite{liu2025step1x} & Open & 1.81 & 0.00 & 1.33 & 0.82 $\pm$ 0.89 & 9.26 $\pm$ 2.59 \\
        \texttt{Qwen-Image-Edit-2509}~\cite{wu2025qwenimage} & Open & 0.60 & 0.49 & 0.67 & 0.62 $\pm$ 0.80 & 10.91 $\pm$ 2.78 \\
        \bottomrule
    \end{tabular}}
\end{table}

\subsection{Main Results}

Table~\ref{tab:main-results} shows that RS-RIE-Bench remains highly challenging for current image editing models. Even the best-performing model, \texttt{gpt-image-2}, achieves only 24.28\% overall accuracy under the strict joint-satisfaction criterion, with a 95\% confidence interval of 20.68--28.28\%. This indicates that existing models are still not able to reliably support reasoning-guided remote sensing image editing.

From the model-type perspective, closed-source models perform better overall than open-source models, but this advantage is only relative rather than decisive. Among the closed-source models, \texttt{gpt-image-2} leads with 24.28\%, followed closely by \texttt{\seqsplit{doubao-seedream-5-0-260128}} at 20.37\%, while \texttt{\seqsplit{gemini-3.1-flash-image-preview}} reaches 12.76\%. In comparison, the strongest open-source model, \texttt{Flux.2-dev}, reaches 6.79\% overall accuracy, and the other open-source models remain clearly behind. Overall, stronger commercial systems do show better editing control and output consistency, but they are still far from reliable remote sensing reasoning-guided editing.

The difficulty also varies across reasoning categories. For \texttt{gpt-image-2} and \texttt{Flux.2-dev}, spatial reasoning reaches 32.00\% and 12.67\%, respectively, which is higher than their performance on temporal reasoning and causal reasoning. This suggests that some geometric adjustment and imaging-constraint tasks are relatively easier. However, this pattern is not universal. For example, the best category of \texttt{\seqsplit{gemini-3.1-flash-image-preview}} is causal reasoning, where it reaches 17.07\%, while its performance on temporal and spatial tasks is weaker. Overall, the three reasoning categories exhibit clear imbalance in difficulty, which indicates that RS-RIE-Bench can effectively expose performance differences across task types.

\begin{figure}[!t]
    \centering
    \includegraphics[width=0.95\textwidth]{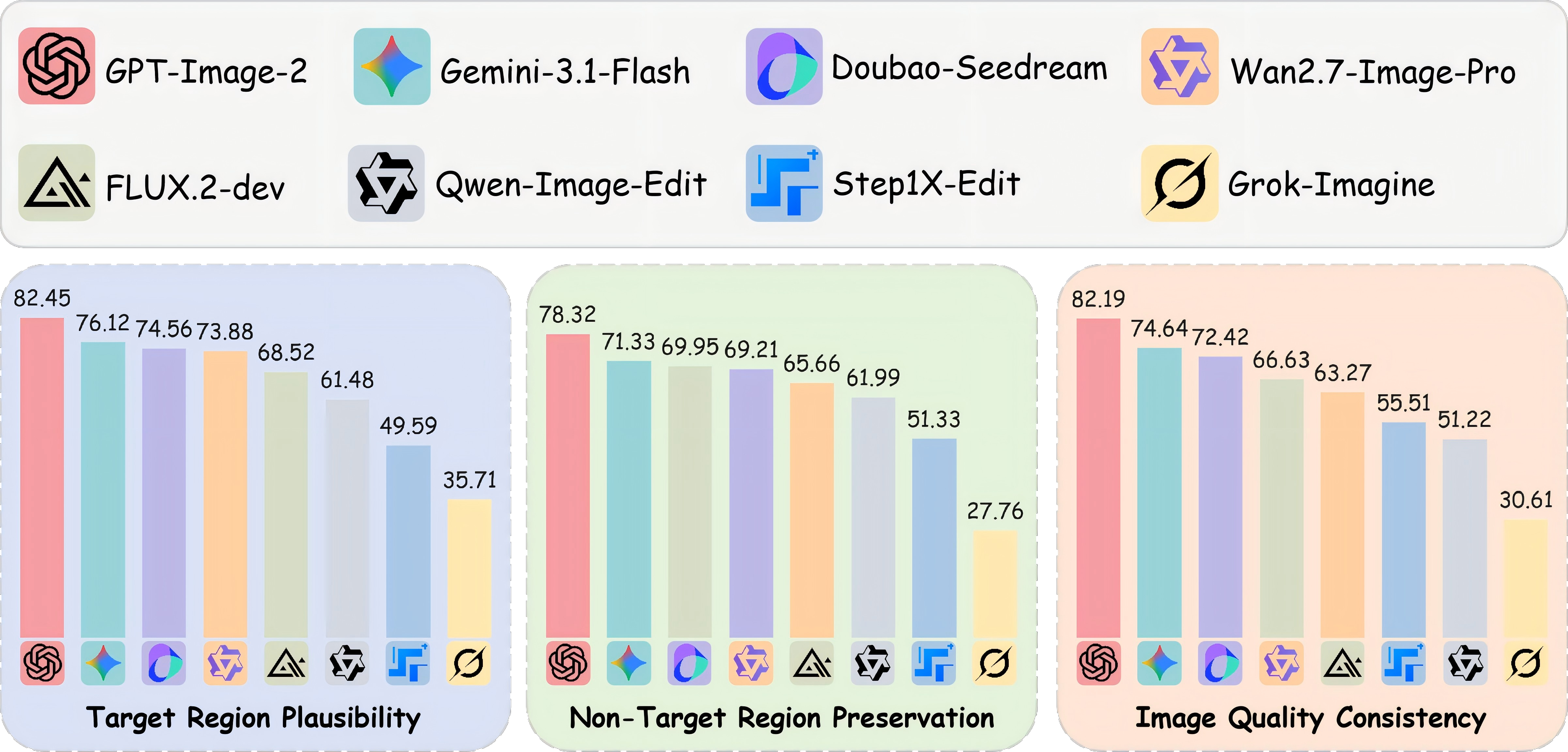}
    \caption{Average normalized performance of different models on the three evaluation dimensions. All scores are normalized to the range 0--100 by \mbox{$(\mathrm{score} - 1) / 4 \times 100$}. Higher values indicate better performance on the corresponding dimension.}
    \label{fig:evaluation-performance}
\end{figure}

To better understand the strengths and limitations of the evaluated models, we further analyze their average performance on the three evaluation dimensions. Figure~\ref{fig:evaluation-performance} provides a more intuitive view of the differences among models on the same tasks and helps explain the numerical gaps reported in the table.

The results show that \texttt{gpt-image-2} has a clear leading advantage on target region plausibility, non-target region preservation, and image quality consistency, which makes it the strongest model on reasoning-guided editing tasks. \texttt{\seqsplit{gemini-3.1-flash-image-preview}} and \texttt{\seqsplit{doubao-seedream-5-0-260128}} form the second tier, with consistently strong performance across the three dimensions, which suggests that they already possess a certain level of remote sensing reasoning-guided editing ability. \texttt{Flux.2-dev} is relatively balanced across the three dimensions, but its overall level still lags behind the leading closed-source models. Although \texttt{\seqsplit{wan2.7-image-pro}} achieves a relatively high score on target region plausibility, it remains clearly weaker on non-target region preservation and image quality consistency. This suggests that the model has some strength in semantic understanding, but still lacks sufficient generation quality.

\begin{figure}[!t]
    \centering
    \includegraphics[width=0.84\textwidth]{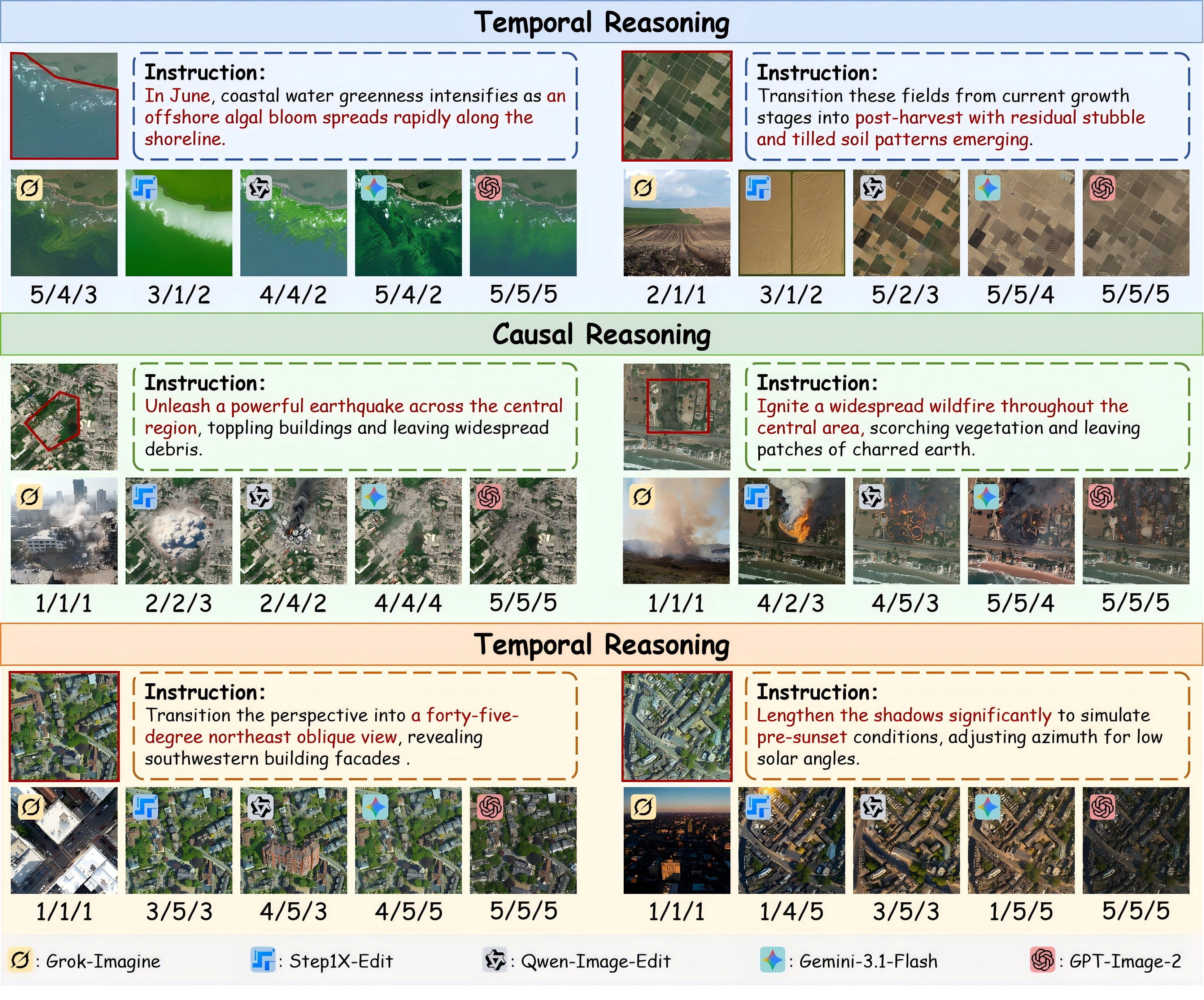}
    \caption{Qualitative comparison of model outputs on representative reasoning-guided remote sensing editing tasks. The score tuple under each output is reported as target region plausibility / non-target region preservation / image quality consistency, where each number ranges from 1 to 5. Different models often satisfy only part of the required constraints, which helps explain the performance differences observed in the quantitative results.}
    \label{fig:model-case}
\end{figure}

The remaining models exhibit more obvious capability gaps. \texttt{Qwen-Image-Edit-2509} and \texttt{Step1X-Edit} reach moderate scores on some dimensions, but they struggle to satisfy all three constraints simultaneously, so their overall accuracy remains close to zero. \texttt{grok-imagine-image} is low across all three dimensions, indicating deficiencies in target editing, region control, and image consistency. As shown in Figure~\ref{fig:model-case}, the score tuple under each output is ordered as target region plausibility, non-target region preservation, and image quality consistency. Many failure cases do involve partial changes, but they only satisfy part of the required constraints and therefore fail under the strict joint-satisfaction criterion.

As a robustness check, we also compute a relaxed joint-4 criterion, under which a sample is counted as successful only when all three dimensions receive at least 4 points. Under this setting, \texttt{gpt-image-2} rises to 61.32\% overall accuracy, and the average success rate across the eight models is 32.23\%, which is still far from a solved benchmark. The relaxed result confirms that the ranking pattern of the strongest models remains broadly stable even when the threshold is loosened.

\subsection{Evaluation Consistency}
\label{sec:evaluation-consistency}

To examine the stability of the automatic evaluation results across different judge models, we compare the scoring consistency of \texttt{gpt-5.1}, \texttt{gemini-3-flash-preview}, and \texttt{qwen-3.5} on the same set of samples. We use quadratically weighted Fleiss' kappa~\cite{fleiss1971measuring,cohen1968weighted} to measure overall consistency on dimension scores. We also report exact agreement on the strict joint-satisfaction decision.

As shown in Table~\ref{tab:judge-consistency}, the three judge models maintain strong overall consistency on dimension scores. The overall $\kappa_w$ values for target region plausibility, non-target region preservation, and image quality consistency are 0.731, 0.726, and 0.707, respectively. This indicates that different judge models usually reach similar conclusions on the plausibility of target edits, the preservation of non-target regions, and the consistency of image quality. Consistency on spatial reasoning is slightly lower than on temporal and causal reasoning. This is understandable because samples involving geometric adjustment and imaging constraints are usually more open-ended and harder to judge.

Under the strict joint-satisfaction decision, the overall exact agreement among the three judges is 75.63\%. This value is lower than the consistency on single-dimension scores, which is expected because disagreement on any one dimension can affect the final success decision. Even so, the agreement rates across task categories remain relatively high. This indicates that the overall conclusions of RS-RIE-Bench are not overly sensitive to the choice of a single judge model.

To further examine whether automatic scores agree with expert judgment, we randomly sample 832 outputs (20.0\%) from the full result set using stratified sampling over all 8 models and 3 task categories; the subset covers 328 causal, 240 spatial, and 264 temporal outputs. The scoring standard is the same as the one used in the MLLM-based evaluation, and the review is independently conducted by four evaluators with remote-sensing or image-interpretation background. We analyze the corresponding expert scores and summarize the human score distribution under each \texttt{gpt-5.1} score level. The results are shown in Table~\ref{tab:human-alignment}.
\begin{table}[!t]
    \centering
    \small
    \caption{Overall consistency of the three judge models on all evaluation samples. $\kappa_w$ denotes quadratically weighted Fleiss' kappa~\cite{fleiss1971measuring,cohen1968weighted} computed from the ordered 1--5 scores. Joint-5 Agreement denotes exact agreement among the three judges on the strict joint-satisfaction decision.}
    \label{tab:judge-consistency}
    \resizebox{\textwidth}{!}{%
    \begin{tabular}{lccccc}
        \toprule
        Task Category & Target $\kappa_w$ & Non-target Region $\kappa_w$ & Quality $\kappa_w$ & Mean $\kappa_w$ & Joint-5 Agreement \\
        \midrule
        Temporal & 0.749 & 0.707 & 0.743 & 0.733 & 77.82 \\
        Causal & 0.710 & 0.720 & 0.681 & 0.704 & 78.45 \\
        Spatial & 0.687 & 0.730 & 0.648 & 0.688 & 69.64 \\
        Overall & 0.731 & 0.726 & 0.707 & 0.722 & 75.63 \\
        \bottomrule
    \end{tabular}}
\end{table}

The distribution shows that expert scores are highly aligned with the model predictions and demonstrate strong overall consistency. When \texttt{gpt-5.1} assigns a score of 1, the corresponding human mean is also usually close to the low-score range. When it assigns a score of 5, the corresponding human means on the three dimensions reach 4.4, 4.8, and 4.4. This suggests that automatic evaluation can reliably identify both obviously failed samples and high-quality outputs. When the model assigns middle scores from 2 to 4, the agreement with human judgment becomes lower. This drop is mainly caused by the subjective nature of the scoring standard. Such subjectivity can also lead to larger differences and potential disagreement even among human experts when they evaluate the same sample. The difference may also come from the model making more fine-grained inspections of generated images and detecting subtle inconsistencies or deviations from the source content that human evaluators may overlook.

From the overall statistics, the MAE values~\cite{willmott2005advantages} for target region plausibility, non-target region preservation, and image quality consistency are 0.546, 0.261, and 0.373, respectively. Under a 1--5 scoring system, all of these values are below 1, which indicates that the absolute deviation between automatic scores and expert scores is generally small. To further measure the global relation between automatic scores and expert scores, we also introduce the Spearman rank correlation coefficient (SRCC)~\cite{spearman1904proof}. The correlations for the three dimensions reach 0.844, 0.905, and 0.868, respectively. Overall, all three dimensions remain well aligned with expert judgment, which indicates that automatic scoring also follows human judgment well at the level of score trends. Taken together, these results suggest that \texttt{gpt-5.1} can serve as the main evaluation tool for RS-RIE-Bench, although expert review is still necessary for highly uncertain cases.

\subsection{Correlation among Evaluation Dimensions}

Finally, we further examine the statistical relations among the three evaluation dimensions to test whether the current evaluation design provides complementary information rather than repeatedly measuring the same ability. Specifically, we compute pairwise Spearman rank correlation coefficients~\cite{spearman1904proof} among target region plausibility, non-target region preservation, and image quality consistency on the full set of automatic scores. We also analyze the correlation between each dimension and the strict joint-satisfaction success indicator.

Table~\ref{tab:dimension-correlation} shows that all three dimensions are positively correlated, but the correlation strength is not the same. This means that the dimensions are not fully independent, but they are also not so strongly correlated that they simply measure the same ability. In particular, the SRCC values between target region plausibility and the other two dimensions are 0.462 and 0.479, both in the moderate range. This shows that whether a target edit follows remote sensing reasoning logic cannot be reduced to whether the non-target region remains stable or whether the image quality looks better. This point is especially important for RS-RIE-Bench, because an edit that appears visually clean may still be physically implausible, while a reasonable target change may still come with non-target-region disturbance or imaging degradation.

The highest correlation appears between non-target region preservation and image quality consistency, with an SRCC of 0.735. This indicates that these two dimensions are statistically more closely related. This result is reasonable because once non-target regions are overly redrawn, more obvious compositing traces, style drift, or local quality degradation often appear at the same time. Even so, this correlation is still not high enough to make the two dimensions interchangeable. Their correlations with the strict joint-satisfaction success indicator are only 0.301 and 0.381, respectively, while target region plausibility also remains similarly important to final success, with a correlation of 0.372. Overall, this analysis supports the current three-dimensional evaluation design of RS-RIE-Bench. The dimensions have meaningful relations, but they are not redundant, so they should still be modeled separately and reported independently.
As a robustness check, we also recompute a relaxed joint-4 criterion, under which a sample is counted as successful only when all three dimensions receive at least 4 points. The overall success rate rises to 32.23\%, and the top three models keep the same order, while the middle and lower ranks change slightly. Across all samples, the mean scores on target region plausibility, non-target region preservation, and image quality consistency are 3.61, 3.47, and 3.48, respectively, which confirms that the benchmark remains challenging even under a looser threshold.

\begin{table}[!t]
    \centering
    \scriptsize
    \setlength{\tabcolsep}{4pt}
    \caption{Calibration analysis between \texttt{gpt-5.1} and expert scores on 832 sampled outputs (20.0\%).} For each score level from 1 to 5 assigned by \texttt{gpt-5.1} on a given dimension, the table reports the sample proportion, human mean, human standard deviation, mean error (human score minus \texttt{gpt-5.1} score), and MAE. The Overall row reports the overall MAE across all samples.
    \label{tab:human-alignment}
    \resizebox{\textwidth}{!}{%
    \begin{tabular}{lcccccccccccccccc}
        \toprule
        \multirow{2}{*}{Judge Score} & \multicolumn{3}{c}{Proportion} & \multicolumn{3}{c}{Human Mean} & \multicolumn{3}{c}{Human Std.} & \multicolumn{3}{c}{Mean Error} & \multicolumn{3}{c}{MAE} \\
        \cmidrule(lr){2-4} \cmidrule(lr){5-7} \cmidrule(lr){8-10} \cmidrule(lr){11-13} \cmidrule(lr){14-16}
        & Target & Non-target & Quality & Target & Non-target & Quality & Target & Non-target & Quality & Target & Non-target & Quality & Target & Non-target & Quality \\
        \midrule
        1 & 21\% & 30\% & 22\% & 1.0 & 1.1 & 1.0 & 0.0 & 0.5 & 0.0 & 0.0 & 0.1 & 0.0 & 0.0 & 0.1 & 0.0 \\
        2 & 14\% & 14\% & 6\% & 1.4 & 2.7 & 1.6 & 0.6 & 1.1 & 0.6 & -0.6 & 0.7 & -0.4 & 0.7 & 0.8 & 0.5 \\
        3 & 8\% & 8\% & 31\% & 2.1 & 3.2 & 2.7 & 1.1 & 0.7 & 0.9 & -0.9 & 0.2 & -0.3 & 1.0 & 0.3 & 0.5 \\
        4 & 33\% & 10\% & 18\% & 3.3 & 4.1 & 3.8 & 1.0 & 0.6 & 0.6 & -0.7 & 0.1 & -0.2 & 0.7 & 0.3 & 0.3 \\
        5 & 23\% & 38\% & 23\% & 4.4 & 4.8 & 4.4 & 1.0 & 0.6 & 0.9 & -0.6 & -0.2 & -0.6 & 0.6 & 0.2 & 0.6 \\
        Overall & -- & -- & -- & -- & -- & -- & -- & -- & -- & -- & -- & -- & 0.5 & 0.3 & 0.4 \\
        \bottomrule
    \end{tabular}}
\end{table}
\begin{table}[!t]
    \centering
    \small
    \caption{Correlation among the three evaluation dimensions. All values are SRCC computed from the full set of evaluation results.}
    \label{tab:dimension-correlation}
    \resizebox{\textwidth}{!}{%
    \begin{tabular}{lcccc}
        \toprule
        Dimension & With Target & With Non-target & With Quality & With Joint-5 Success \\
        \midrule
        Target Region Plausibility & -- & 0.462 & 0.479 & 0.372 \\
        Non-target Region Preservation & 0.462 & -- & 0.735 & 0.301 \\
        Image Quality Consistency & 0.479 & 0.735 & -- & 0.381 \\
        \bottomrule
    \end{tabular}}
\end{table}
\section{Conclusion}
This study has several limitations. First, RS-RIE-Bench focuses on RGB optical imagery and therefore evaluates sensor consistency through visual and texture-level proxies rather than calibrated multispectral radiometry. Second, several source datasets impose redistribution constraints, so the benchmark release separates metadata and reconstruction scripts from restricted imagery. Third, MLLM-based evaluation remains a scalable proxy for expert assessment and should be interpreted together with human validation and objective preservation metrics. Finally, the current benchmark size is suitable for diagnostic evaluation but should be expanded with hidden splits, additional sensors, and broader geographic coverage in future versions.

\bibliographystyle{plainnat}
\bibliography{references}

@inproceedings{brooks2023instructpix2pix,
    author    = {Brooks, Tim and Holynski, Aleksander and Efros, Alexei A.},
    title     = {InstructPix2Pix: Learning To Follow Image Editing Instructions},
    booktitle = {Proceedings of the IEEE/CVF Conference on Computer Vision and Pattern Recognition (CVPR)},
    month     = {June},
    year      = {2023},
    pages     = {18392-18402}
}

@inproceedings{wang2023editbench,
    author    = {Wang, Su and Saharia, Chitwan and Montgomery, Ceslee and Pont-Tuset, Jordi and Noy, Shai and Pellegrini, Stefano and Onoe, Yasumasa and Laszlo, Sarah and Fleet, David J. and Soricut, Radu and Baldridge, Jason and Norouzi, Mohammad and Anderson, Peter and Chan, William},
    title     = {Imagen Editor and EditBench: Advancing and Evaluating Text-Guided Image Inpainting},
    booktitle = {Proceedings of the IEEE/CVF Conference on Computer Vision and Pattern Recognition (CVPR)},
    month     = {June},
    year      = {2023},
    pages     = {18359-18369}
}

@inproceedings{kuckreja2024geochat,
    author    = {Kuckreja, Kartik and Danish, Muhammad Sohail and Naseer, Muzammal and Das, Abhijit and Khan, Salman and Khan, Fahad Shahbaz},
    title     = {GeoChat: Grounded Large Vision-Language Model for Remote Sensing},
    booktitle = {Proceedings of the IEEE/CVF Conference on Computer Vision and Pattern Recognition (CVPR)},
    month     = {June},
    year      = {2024},
    pages     = {27831-27840}
}

@article{luo2026vlrs,
      title={VLRS-Bench: A Vision-Language Reasoning Benchmark for Remote Sensing}, 
      author={Zhiming Luo and Di Wang and Haonan Guo and Jing Zhang and Bo Du},
      year={2026},
      eprint={2602.07045},
      archivePrefix={arXiv},
      primaryClass={cs.CV},
      url={https://arxiv.org/abs/2602.07045}, 
}

@article{chen2026rsedit,
  author={Chen, Zhenyuan and Zhang, Zechuan and Zhang, Feng},
  journal={IEEE Geoscience and Remote Sensing Letters}, 
  title={RSEdit: Text-Guided Image Editing for Remote Sensing}, 
  year={2026},
  volume={23},
  number={},
  pages={6011905-6011905},
  doi={10.1109/LGRS.2026.3695484}}

@inproceedings{li2023blip2,
  title = 	 {{BLIP}-2: Bootstrapping Language-Image Pre-training with Frozen Image Encoders and Large Language Models},
  author =       {Li, Junnan and Li, Dongxu and Savarese, Silvio and Hoi, Steven},
  booktitle = 	 {Proceedings of the 40th International Conference on Machine Learning},
  pages = 	 {19730--19742},
  year = 	 {2023},
  editor = 	 {Krause, Andreas and Brunskill, Emma and Cho, Kyunghyun and Engelhardt, Barbara and Sabato, Sivan and Scarlett, Jonathan},
  volume = 	 {202},
  series = 	 {Proceedings of Machine Learning Research},
  month = 	 {23--29 Jul},
  publisher =    {PMLR},
  url = 	 {https://proceedings.mlr.press/v202/li23q.html}
}

@article{dai2023instructblip,
  author = {Dai, Wenliang and Li, Junnan and Li, Dongxu and Tiong, Anthony and Zhao, Junqi and Wang, Weisheng and Sebe, Nicu and Hoi, Steven C. H.},
  title = {InstructBLIP: Towards General-Purpose Vision-Language Models with Instruction Tuning},
  booktitle = {Advances in Neural Information Processing Systems},
  editor = {A. Oh and T. Naumann and A. Globerson and K. Saenko and M. Hardt and S. Levine},
  pages = {49250--49267},
  publisher = {Curran Associates, Inc.},
  volume = {36},
  year = {2023},
  url = {https://proceedings.neurips.cc/paper_files/paper/2023/file/9a6a435e75419a836fe47ab6793623e6-Paper-Conference.pdf}
}

@inproceedings{zhang2023magicbrush,
 author = {Zhang, Kai and Mo, Lingbo and Chen, Wenhu and Sun, Huan and Su, Yu},
 booktitle = {Advances in Neural Information Processing Systems},
 editor = {A. Oh and T. Naumann and A. Globerson and K. Saenko and M. Hardt and S. Levine},
 pages = {31428--31449},
 publisher = {Curran Associates, Inc.},
 title = {MagicBrush: A Manually Annotated Dataset for Instruction-Guided Image Editing},
 url = {https://proceedings.neurips.cc/paper_files/paper/2023/file/64008fa30cba9b4d1ab1bd3bd3d57d61-Paper-Datasets_and_Benchmarks.pdf},
 volume = {36},
 year = {2023}
}

@article{zhao2026risebench,
 author = {Zhao, Xiangyu and Zhang, Peiyuan and Tang, Kexian and Zhu, Xiaorong and Li, Hao and Chai, Wenhao and Zhang, Zicheng and Xia, Renqiu and Zhai, Guangtao and Yan, Junchi and Yang, Hua and Yang, Xue and Duan, Haodong},
 booktitle = {Advances in Neural Information Processing Systems},
 editor = {D. Belgrave and C. Zhang and H. Lin and R. Pascanu and P. Koniusz and M. Ghassemi and N. Chen},
 pages = {},
 publisher = {Curran Associates, Inc.},
 title = {Envisioning Beyond the Pixels: Benchmarking Reasoning-Informed Visual Editing},
 url = {https://proceedings.neurips.cc/paper_files/paper/2025/file/fe79898dcf078ec54b6feeea10ebb751-Paper-Datasets_and_Benchmarks_Track.pdf},
 volume = {38},
 year = {2025}
}

@article{liu2026grade,
      title={GRADE: Benchmarking Discipline-Informed Reasoning in Image Editing}, 
      author={Mingxin Liu and Ziqian Fan and Zhaokai Wang and Leyao Gu and Zirun Zhu and Yiguo He and Yuchen Yang and Changyao Tian and Xiangyu Zhao and Ning Liao and Shaofeng Zhang and Qibing Ren and Zhihang Zhong and Xuanhe Zhou and Junchi Yan and Xue Yang},
      year={2026},
      eprint={2603.12264},
      archivePrefix={arXiv},
      primaryClass={cs.CV},
      url={https://arxiv.org/abs/2603.12264}, 
}

@article{wu2026krisbench,
 author = {Wu, Yongliang and Li, Zonghui and Hu, Xinting and Ye, Xinyu and Zeng, Xianfang and Yu, Gang and Zhu, Wenbo and Schiele, Bernt and Yang, Ming-Hsuan and Yang, Xu},
 booktitle = {Advances in Neural Information Processing Systems},
 editor = {D. Belgrave and C. Zhang and H. Lin and R. Pascanu and P. Koniusz and M. Ghassemi and N. Chen},
 pages = {},
 publisher = {Curran Associates, Inc.},
 title = {KRIS-Bench: Benchmarking Next-Level Intelligent Image Editing Models},
 url = {https://proceedings.neurips.cc/paper_files/paper/2025/file/e619b285582fb12f4c3de3a507b8b99c-Paper-Datasets_and_Benchmarks_Track.pdf},
 volume = {38},
 year = {2025}
}

@article{wang2026disasterm3,
 author = {Wang, Junjue and Xuan, Weihao and Qi, Heli and Liu, Zhihao and Liu, Kunyi and Wu, Yuhan and Chen, Hongruixuan and SONG, JIAN and Xia, Junshi and Zheng, Zhuo and YOKOYA, Naoto},
 booktitle = {Advances in Neural Information Processing Systems},
 editor = {D. Belgrave and C. Zhang and H. Lin and R. Pascanu and P. Koniusz and M. Ghassemi and N. Chen},
 pages = {},
 publisher = {Curran Associates, Inc.},
 title = {DisasterM3: A Remote Sensing Vision-Language Dataset for Disaster Damage Assessment and Response},
  url = {https://proceedings.neurips.cc/paper_files/paper/2025/file/ec80d18205e39d42a27192d5f3ddd688-Paper-Datasets_and_Benchmarks_Track.pdf},
  volume = {38},
  year = {2025}
}

@inproceedings{rombach2022ldm,
    author    = {Rombach, Robin and Blattmann, Andreas and Lorenz, Dominik and Esser, Patrick and Ommer, Bj\"orn},
    title     = {High-Resolution Image Synthesis With Latent Diffusion Models},
    booktitle = {Proceedings of the IEEE/CVF Conference on Computer Vision and Pattern Recognition (CVPR)},
    month     = {June},
    year      = {2022},
    pages     = {10684-10695}
}

@inproceedings{esser2024sd3,
title={Scaling Rectified Flow Transformers for High-Resolution Image Synthesis},
author={Patrick Esser and Sumith Kulal and Andreas Blattmann and Rahim Entezari and Jonas M{\"u}ller and Harry Saini and Yam Levi and Dominik Lorenz and Axel Sauer and Frederic Boesel and Dustin Podell and Tim Dockhorn and Zion English and Robin Rombach},
booktitle={Forty-first International Conference on Machine Learning},
year={2024},
url={https://openreview.net/forum?id=FPnUhsQJ5B}
}

@article{labs2025flux,
  title={{FLUX}.2: Frontier Visual Intelligence},
  author={{Black Forest Labs}},
  year={2025},
  url={https://blackforestlabs.ai/announcements/},
  note={Official Black Forest Labs announcement page for the FLUX.2 model family}
}

@article{openai2025gpt5,
  title={OpenAI {GPT}-5 System Card},
  author={{OpenAI}},
  journal={arXiv preprint arXiv:2601.03267},
  year={2026}
}

@article{comanici2025gemini25,
  title={Gemini 2.5: Pushing the Frontier with Advanced Reasoning, Multimodality, Long Context, and Next Generation Agentic Capabilities},
  author={Comanici, Gabriel and Bieber, Emily and Schaekermann, Mike and others},
  journal={arXiv preprint arXiv:2507.06261},
  year={2025}
}

@article{deng2025bagel,
  title={Emerging Properties in Unified Multimodal Pretraining},
  author={Deng, Chaorui and Zhu, Deyao and Li, Kaixuan and others},
  journal={arXiv preprint arXiv:2505.14683},
  year={2025}
}

@inproceedings{xiao2025omnigen,
  title={OmniGen: Unified Image Generation},
  author={Xiao, Shitao and Wang, Yukang and Zhou, Junjie and others},
  booktitle={Proceedings of the IEEE/CVF Conference on Computer Vision and Pattern Recognition},
  pages={13294--13304},
  year={2025}
}

@article{wu2025qwenimage,
  title={Qwen-Image-Edit-2509: Multi-Image Support, Improved Consistency},
  author={{Qwen Team}},
  year={2025},
  url={https://qwen.ai/blog?from=research.latest-advancements-list&id=7a90090115ee193ce6a7f619522771dd9696dd93},
  note={Official Qwen blog page for \texttt{Qwen-Image-Edit-2509}}
}

@article{liu2025step1x,
  title={Step1X-Edit: A Practical Framework for General Image Editing},
  author={Liu, Shilong and Han, Yifan and Xing, Peng and others},
  journal={arXiv preprint arXiv:2504.17761},
  year={2025}
}

@inproceedings{zheng2023changen,
  title={Scalable Multi-Temporal Remote Sensing Change Data Generation via Simulating Stochastic Change Process},
  author={Zheng, Zhuo and Tian, Shiqi and Ma, Ailong and Zhang, Liangpei and Zhong, Yanfei},
  booktitle={Proceedings of the IEEE/CVF International Conference on Computer Vision},
  pages={21818--21827},
  year={2023}
}

@article{zheng2023judging,
 author = {Zheng, Lianmin and Chiang, Wei-Lin and Sheng, Ying and Zhuang, Siyuan and Wu, Zhanghao and Zhuang, Yonghao and Lin, Zi and Li, Zhuohan and Li, Dacheng and Xing, Eric and Zhang, Hao and Gonzalez, Joseph and Stoica, Ion},
 booktitle = {Advances in Neural Information Processing Systems},
 editor = {A. Oh and T. Naumann and A. Globerson and K. Saenko and M. Hardt and S. Levine},
 pages = {46595--46623},
 publisher = {Curran Associates, Inc.},
 title = {Judging LLM-as-a-Judge with MT-Bench and Chatbot Arena},
 url = {https://proceedings.neurips.cc/paper_files/paper/2023/file/91f18a1287b398d378ef22505bf41832-Paper-Datasets_and_Benchmarks.pdf},
 volume = {36},
 year = {2023}
}

@inproceedings{xia2018dota,
author = {Xia, Gui-Song and Bai, Xiang and Ding, Jian and Zhu, Zhen and Belongie, Serge and Luo, Jiebo and Datcu, Mihai and Pelillo, Marcello and Zhang, Liangpei},
title = {DOTA: A Large-Scale Dataset for Object Detection in Aerial Images},
booktitle = {Proceedings of the IEEE Conference on Computer Vision and Pattern Recognition (CVPR)},
month = {June},
year = {2018}
}

@article{zhou2018patternnet,
title = {PatternNet: A benchmark dataset for performance evaluation of remote sensing image retrieval},
journal = {ISPRS Journal of Photogrammetry and Remote Sensing},
volume = {145},
pages = {197-209},
year = {2018},
note = {Deep Learning RS Data},
issn = {0924-2716},
doi = {https://doi.org/10.1016/j.isprsjprs.2018.01.004},
url = {https://www.sciencedirect.com/science/article/pii/S0924271618300042},
author = {Weixun Zhou and Shawn Newsam and Congmin Li and Zhenfeng Shao}
}

@article{cheng2017nwpu,
  author={Cheng, Gong and Han, Junwei and Lu, Xiaoqiang},
  journal={Proceedings of the IEEE}, 
  title={Remote Sensing Image Scene Classification: Benchmark and State of the Art}, 
  year={2017},
  volume={105},
  number={10},
  pages={1865-1883},
  doi={10.1109/JPROC.2017.2675998}}

@article{amirkolaee2024adatreeformer,
title = {AdaTreeFormer: Few shot domain adaptation for tree counting from a single high-resolution image},
journal = {ISPRS Journal of Photogrammetry and Remote Sensing},
volume = {214},
pages = {193-208},
year = {2024},
issn = {0924-2716},
doi = {https://doi.org/10.1016/j.isprsjprs.2024.06.015},
url = {https://www.sciencedirect.com/science/article/pii/S0924271624002533},
author = {Hamed Amini Amirkolaee and Miaojing Shi and Lianghua He and Mark Mulligan}
}

@article{gupta2019xbd,
author = {Gupta, Ritwik and Goodman, Bryce and Patel, Nirav and Hosfelt, Ricky and Sajeev, Sandra and Heim, Eric and Doshi, Jigar and Lucas, Keane and Choset, Howie and Gaston, Matthew},
title = {Creating xBD: A Dataset for Assessing Building Damage from Satellite Imagery},
booktitle = {Proceedings of the IEEE/CVF Conference on Computer Vision and Pattern Recognition (CVPR) Workshops},
month = {June},
year = {2019}
}

@article{openai2024gpt4o,
  title={{GPT}-4o System Card},
  author={{OpenAI}},
  journal={arXiv preprint arXiv:2410.21276},
  year={2024}
}

@article{openai2024gpt51,
  title={OpenAI {GPT}-5 System Card},
  author={{OpenAI}},
  journal={arXiv preprint arXiv:2601.03267},
  year={2026},
  note={Official OpenAI system card used for \texttt{gpt-5.1}}
}

@misc{openai2026gptimage2,
  title={{GPT} Image 2 Model},
  author={{OpenAI}},
  year={2026},
  url={https://developers.openai.com/api/docs/models/gpt-image-2},
  note={Official OpenAI API model documentation for \texttt{gpt-image-2}}
}

@article{google2026gemini31flashimage,
  title={Gemini 3.1 Flash Image},
  author={{Google}},
  year={2026},
  url={https://ai.google.dev/gemini-api/docs/models/gemini-3.1-flash-image},
  note={Official Google AI for Developers model page for \texttt{gemini-3.1-flash-image-preview}}
}

@article{mao2026wanimage,
  title={Wan 2.7 Pro},
  author={{Wan AI}},
  year={2026},
  url={https://create.wan.video/explore/image/generate?model=wan2.7-pro},
  note={Official Wan AI product page for \texttt{wan2.7-image-pro}}
}

@misc{xai2026grokimagineimage,
  title={Imagine API: Generate Videos, Images, and Audio},
  author={{xAI}},
  year={2026},
  url={https://x.ai/api/imagine?mode=image},
  note={Official xAI API documentation for \texttt{grok-imagine-image}}
}

@article{gao2025seedream3,
  title={Seedream 3.0 Text-to-Image Model Technical Report Released},
  author={{ByteDance Seed}},
  year={2025},
  url={https://seed.bytedance.com/en/public_papers/seedream-3-0-technical-report?view_from=research},
  note={Official ByteDance Seed release page for \texttt{doubao-seedream-5-0-260128}}
}

@article{google2026gemini3flashpreview,
  title={Gemini 3 Flash Preview},
  author={{Google}},
  year={2026},
  url={https://ai.google.dev/gemini-api/docs/models/gemini-3-flash-preview},
  note={Official Google AI for Developers model page for \texttt{gemini-3-flash-preview}}
}

@article{qwen2026qwen35omni,
  title={Qwen3.5-Omni: Scaling Up, Toward Native Omni-Modal AGI},
  author={{Qwen Team}},
  year={2026},
  url={https://qwen.ai/blog?id=qwen3.5-omni},
  note={Official Qwen blog page for \texttt{qwen-3.5}}
}

@article{wilson1927probable,
author = {Wilson, Edwin B.},
title = {Probable Inference, the Law of Succession, and Statistical Inference},
journal = {Journal of the American Statistical Association},
volume = {22},
number = {158},
pages = {209--212},
year = {1927},
publisher = {Taylor \& Francis},
doi = {10.1080/01621459.1927.10502953}
}

@article{fleiss1971measuring,
  title={Measuring Nominal Scale Agreement among Many Raters},
  author={Fleiss, Joseph L.},
  journal={Psychological Bulletin},
  volume={76},
  number={5},
  pages={378--382},
  year={1971}
}

@article{cohen1968weighted,
  title={Weighted Kappa: Nominal Scale Agreement with Provision for Scaled Disagreement or Partial Credit},
  author={Cohen, Jacob},
  journal={Psychological Bulletin},
  volume={70},
  number={4},
  pages={213--220},
  year={1968}
}

@article{willmott2005advantages,
  author  = {Willmott, C. J. and Matsuura, K.},
  title   = {Advantages of the mean absolute error ({MAE}) over the root mean square error ({RMSE}) in assessing average model performance},
  journal = {Climate Research},
  year    = {2005},
  volume  = {30},
  pages   = {79--82},
  doi     = {10.3354/cr030079}
}

@article{spearman1904proof,
  title={The proof and measurement of association between two things.},
  author={Spearman, Charles},
  year={1961},
  publisher={Appleton-Century-Crofts}
}

\appendix
\section{Full Model Settings}

This appendix summarizes the model settings, instruction annotation workflow, and evaluation protocol used in RS-RIE-Bench. To reduce scale-related bias, each model is run once per sample without repeated sampling, and the original remote sensing image is pre-cropped or resized to match the model's default output resolution.

\subsection{Model List}

Table A1 summarizes all image editing models evaluated in this paper, including the model name, model type, access mode, input resolution, and output resolution.

\begin{table}[H]
    \centering
    \small
    \caption{Image editing model settings used in RS-RIE-Bench.}
    \label{tab:app-model-settings}
    \resizebox{\textwidth}{!}{%
    \begin{tabular}{lcccc}
        \toprule
        Model Name & Model Type & Access Mode & Input Resolution & Output Resolution \\
        \midrule
        gpt-image-2~\cite{openai2026gptimage2} & Closed-source & API & 1254$\times$1254 & 1254$\times$1254 \\
        gemini-3.1-flash-image-preview~\cite{google2026gemini31flashimage} & Closed-source & API & 1024$\times$1024 & 1024$\times$1024 \\
        wan2.7-image-pro~\cite{mao2026wanimage} & Closed-source & API & 2048$\times$2048 & 2048$\times$2048 \\
        grok-imagine-image~\cite{xai2026grokimagineimage} & Closed-source & API & 1024$\times$1024 & 1024$\times$1024 + 1168$\times$784 (failed samples) \\
        doubao-seedream-5-0-260128~\cite{gao2025seedream3} & Closed-source & API & 2048$\times$2048 & 2048$\times$2048 \\
        Qwen-Image-Edit-2509~\cite{wu2025qwenimage} & Open-source & API & 1024$\times$1024 & 1024$\times$1024 \\
        Flux.2-dev~\cite{labs2025flux} & Open-source & Local inference & 1024$\times$1024 / 256$\times$256 & 1024$\times$1024 / 256$\times$256 \\
        Step1X-Edit~\cite{liu2025step1x} & Open-source & Local inference & 1024$\times$1024 / 256$\times$256 & 1024$\times$1024 / 256$\times$256 \\
        \bottomrule
    \end{tabular}}
\end{table}

\subsection{Filtering Funnel}

Table~\ref{tab:construction-funnel} summarizes the benchmark construction funnel. We begin with 521 pre-screened RGB optical remote sensing candidates. After quality screening, 516 samples remain; taxonomy assignment keeps 505 cases; human review shortlists 494 samples; and the final benchmark contains 486 images.

\begin{table}[H]
    \centering
    \small
    \caption{Filtering funnel for RS-RIE-Bench construction.}
    \label{tab:construction-funnel}
    \begin{tabularx}{\textwidth}{l c X}
        \toprule
        Stage & Count & Note \\
        \midrule
        Candidate collection & 521 & Pre-screened RGB optical remote sensing images collected from the raw pool \\
        Quality screening & 516 & Images failing basic visual clarity or annotation completeness checks are removed \\
        Taxonomy assignment & 505 & Ambiguous or multi-label cases are excluded during task-category assignment \\
        Human review & 494 & \texttt{gpt-4o}~\cite{openai2024gpt4o}, \texttt{gemini-2.5}~\cite{comanici2025gemini25}, and human experts independently review the remaining samples \\
        Final retention & 486 & Final benchmark size \\
        \bottomrule
    \end{tabularx}
\end{table}

\subsection{Unified Generation Protocol}

All image editing outputs are generated under a single-sample protocol. Each model follows its default output resolution, and the input image is pre-cropped or resized to match that scale. For grok-imagine-image, some samples produce outputs at 1168$\times$784; we record these separately. No extra post-processing is applied.

\subsection{Instruction Annotation and Review}

We manually write the English prompts for the three task families and then review them with \texttt{gpt-4o}~\cite{openai2024gpt4o} and \texttt{gemini-2.5}~\cite{comanici2025gemini25} for ambiguity, category consistency, and output-format compliance. Temporal and spatial tasks use single-temporal remote sensing images, while causal tasks use paired pre-disaster and post-disaster images. All pages require a single-sentence English output and append the same background-preserving suffix.

Temporal instruction annotation uses a single-temporal remote sensing image. The manually written prompt asks for about 20 English words, a single sentence, and coverage of seasonal change, algae bloom, leaf turning, lake drying, swamp drying, port holiday, rush hour, red-eye airport, beach vacation, station crowding, glacier melt, sea-ice melt, parking-lot abandonment, farmland desertification, lake disappearance, and desert afforestation. The output must append the background-preserving suffix.

Spatial instruction annotation also uses a single-temporal remote sensing image. The manually written prompt requires one English sentence and covers viewpoint change, lighting change, object relocation, and resolution scaling. The viewpoint branch further constrains the verb, tilt angle, and direction, while the object-relocation branch further constrains unique target selection. No layout explanation is allowed, and the background-preserving suffix must be appended.

Causal instruction annotation uses paired pre-disaster and post-disaster images. The manually written prompt reverse-describes a single-sentence English disaster simulation instruction from the image pair, and the sentence must include spatial location, destructive action, and final damage state. It covers earthquake, flood, wildfire, and wind scenarios, and also appends the background-preserving suffix.

\subsection{Full Prompt Text}

For reproducibility, we provide representative prompt segments for the three task families. Each family follows the same organization: role setting, task description, output constraints, and examples.

\subsubsection{Temporal Reasoning Prompt}

The temporal reasoning page is organized as a task overview, a subtask constraint table, and a representative prompt example. The subtasks cover algal bloom, vegetation phenology, lake recession, wetland drying, port holiday closure, road rush hour, night airport operation, beach tourism peak, rail station crowding, glacier retreat, sea-ice melt, abandoned parking lots, farmland desertification, lake disappearance, and desert afforestation.

\begin{table}[H]
    \centering
    \small
    \caption{Temporal reasoning subtasks and output constraints.}
    \label{tab:app-temporal-subtasks}
    \resizebox{\textwidth}{!}{%
    \begin{tabular}{lccc}
        \toprule
        Subtask & Task Goal & Core Constraint & Output Requirement \\
        \midrule
        algae\_bloom & Offshore algal or phytoplankton bloom & Focus on marine ecology and sea-surface color change & About 20 words, single sentence \\
        leaf\_turning & Vegetation phenology transition & Focus on yellow/red leaf change and chlorophyll decay & About 20 words, single sentence \\
        lake\_drying\_season & Seasonal lake recession & Focus on shoreline retreat and exposed lake bed & About 20 words, single sentence \\
        swamp\_drying & Wetland drying & Focus on water-level decline and exposed mudland & About 20 words, single sentence \\
        port\_holiday & Port holiday closure & Focus on berth usage and shipping rhythm & About 20 words, single sentence \\
        road\_rush\_hour & Road rush hour & Focus on vehicle density increase & About 20 words, single sentence \\
        red\_eye\_airport & Night airport operation & Focus on airport lighting and aircraft layout changes & About 20 words, single sentence \\
        beach\_vacation & Beach tourism peak & Focus on crowd-density change & About 20 words, single sentence \\
        station\_golden\_week & Rail station crowd peak & Focus on track occupancy and train density & About 20 words, single sentence \\
        glacier\_melt & Long-term glacier melt & Focus on multi-decade retreat and exposed bedrock & About 25 words, include a time phrase \\
        sea\_ice\_melt & Long-term sea-ice melt & Focus on multi-year fragmentation and open water exposure & About 20 words, include a time phrase \\
        parkinglot\_abandoned & Long-term parking-lot abandonment & Focus on infrastructure decay and vegetation invasion & About 20 words, single sentence \\
        farmland\_desert & Farmland desertification & Focus on farmland evolving into arid sand & About 25 words, include a time phrase \\
        lake\_disappear & Long-term lake disappearance & Focus on complete drying and salt-crust formation & About 20 words, single sentence \\
        desert\_afforestation & Desert afforestation & Focus on ecological restoration and windbreak structure & About 25 words, include a time phrase \\
        \bottomrule
    \end{tabular}}
\end{table}

\begin{Verbatim}[fontsize=\scriptsize,frame=single,rulecolor=\color{black},framesep=2mm,baselinestretch=0.95,breaklines=true,breakanywhere=true,commandchars=\\\{\}]
You are an expert in marine remote sensing and oceanography. Diagnose the current
ocean surface state in the satellite image, and reason a sudden offshore algal bloom
event.
Write a single instruction in English. Focus strictly on marine ecology (e.g., massive
green macroalgal or phytoplankton outbreaks covering coastal waters). Keep the
sentence around 20 words. Use structure: [Action Verb] + [Transformation of sea
surface and coastal water colors]. No preamble, no quotes.

Example:
- Input Image context: High-resolution remote sensing image during mid-winter
  (January-February) over a large offshore sea or semi-closed gulf, where the water
  surface presents a deep, homogeneous, low-reflective pure dark blue or black
  texture.
- Expected Output Instruction (around 20 words): Simulate a massive spring
  phytoplankton bloom expanding across this offshore sea area during April and May.
\end{Verbatim}

\subsubsection{Spatial Reasoning Prompt}

The spatial reasoning page contains four branches: nadir-to-oblique viewpoint change, shadow and pre-sunset illumination change, aircraft relocation, and single-building rotation with texture completion. All branches require a single English sentence without any explanatory prefix. For the viewpoint branch, the model must use a 30/35/40/45-degree tilt toward a northeast, southeast, northwest, or southwest direction; the verb is restricted to Tilt, Shift, Transform, or Transition; and the answer must stay within 30 words while including both angle and direction. For shadow inversion, the verb is restricted to Lengthen, Extend, Transform, or Shift, and the answer should focus on shadow geometry and pre-sunset illumination in about 20 words. For aircraft relocation, the prompt asks the model to identify the unique target aircraft by color, size extrema, and direction extrema before describing the destination, with no layout explanation. For building rotation, the prompt randomly chooses a 90- or 180-degree rotation and requires completion of the surrounding surface texture, ending with environmental texture completion.

\begin{Verbatim}[fontsize=\scriptsize,frame=single,rulecolor=\color{black},framesep=2mm,baselinestretch=0.95,breaklines=true,breakanywhere=true,commandchars=\\\{\}]
You are an expert in geometric remote sensing, photogrammetry, and 3D urban spatial
analysis. Diagnose the current nadir perspective of high-rise structures in the
satellite image, and reason its NEXT logical transformation to an oblique viewpoint.
Write a single instruction in English. Focus strictly on geometric distortion effects.
You MUST START the instruction sentence strictly with the active verb
"${randomV1Verb}". You MUST strictly use exactly "${randomDegree.text}-degree" as the
tilt angle, and strictly use "${randomDirection.dir}" as the target direction for this
task. Keep the sentence strictly within 30 words. Use structure: [${randomV1Verb}] +
[Transformation of perspective and structural geometry with exactly
"${randomDegree.text}-degree" and direction "${randomDirection.dir}"]. Do NOT mention
background details or layout status. No preamble, no quotes.

Example:
- Expected Output Instruction: ${randomV1Verb} the perspective into a
  ${randomDegree.text}-degree ${randomDirection.dir} off-nadir view, exposing
  ${randomDirection.expose} building facades while geometrically occluding adjacent
  ${randomDirection.occlude} streets.
\end{Verbatim}

\subsubsection{Causal Reasoning Prompt}

The causal reasoning page uses a paired-temporal-input and single-sentence-output structure. The key difference from the temporal page is that causal prompts require the model to reverse-engineer a disaster event from the before-and-after images and explicitly describe the spatial location, destructive action, and final damage state.

\begin{table}[H]
    \centering
    \small
    \caption{Causal reasoning task requirements.}
    \label{tab:app-causal-requirements}
    \resizebox{\textwidth}{!}{%
    \begin{tabular}{lccc}
        \toprule
        Step & Task Goal & Core Constraint & Output Requirement \\
        \midrule
        Reverse inference & Identify the disaster type from paired images & Must infer flood, wildfire, earthquake, or wind mechanisms & Single English sentence \\
        Spatial location & Specify the affected spatial extent & Must include the spatial location phrase & Single English sentence \\
        Destructive action & Describe the concrete disaster action & Must include the destructive action phrase & Single English sentence \\
        Final state & Describe the final damage state & Must include the final damage-state phrase & Single English sentence \\
        Suffix constraint & Preserve the original image structure & Must append the required background-preserving suffix & No explanatory text \\
        \bottomrule
    \end{tabular}}
\end{table}

\begin{Verbatim}[fontsize=\scriptsize,frame=single,rulecolor=\color{black},framesep=2mm,baselinestretch=0.95,breaklines=true,breakanywhere=true,commandchars=\\\{\}]
You are a geospatial disaster assessment expert. I have provided you with two
satellite images of the same area: Time A (pre-disaster pristine state) and Time B
(post-disaster damaged state).

Please observe the physical destruction between the two images, ignoring minor
lighting/sensor differences. Focus on identifying areas that suffered from disasters
(e.g., collapsed buildings, flooded streets, scorched earth). Reverse-engineer the
event and write a single [Disaster Simulation Instruction] with a sense of destructive
action, guiding me on how to transform the environment from Time A to Time B.

Requirements:
1. Identify the likely destructive force (flood, wildfire, earthquake, wind).
2. MUST include: [Spatial Location] + [Destructive Action] + [Final Damage State].
3. Output ONLY a single-sentence instruction. No explanations, no quotes.
4. You MUST append this exact phrase at the very end of your generated instruction:
   " Keeping the original image resolution, overall geographic layout, and
   background details strictly unchanged."

Examples:
- Sweep a severe wildfire across the residential zone in the upper right, burning
  the vegetation and reducing the buildings to ash. Keeping the original image
  resolution, overall geographic layout, and background details strictly unchanged.
- Inundate the riverbanks in the lower half with heavy flooding, submerging the
  roads under muddy water. Keeping the original image resolution, overall geographic
  layout, and background details strictly unchanged.
\end{Verbatim}

\section{Full MLLM-as-a-Judge Protocol}

This appendix describes the MLLM-as-a-Judge~\cite{zheng2023judging} protocol used in automatic evaluation. We treat the judge as a scalable proxy rather than ground truth and use a three-dimensional scoring framework aligned with human validation. Temporal, causal, and spatial tasks each use task-specific prompts, but the core rubric and the 1-5 scale remain the same. Human validation uses the same task prompt and score scale.

\subsection{Judge Model and Overall Setup}

Automatic evaluation uses MLLM-as-a-Judge~\cite{zheng2023judging} to score each output along three dimensions. The primary judge is \texttt{gpt-5.1}~\cite{openai2024gpt51}; the comparison judges are \texttt{gemini-3-flash-preview}~\cite{google2026gemini3flashpreview} and \texttt{qwen-3.5}~\cite{qwen2026qwen35omni}. The input includes the original image, the generated image, and the editing instruction in a single-turn prompt. The judge output must include a \texttt{Reason:} explanation and a \texttt{Final Score:} conclusion. We set \texttt{temperature = 0.0}, and outputs without an extractable \texttt{Final Score} are marked as \texttt{Format Error}.

\begin{table}[H]
    \centering
    \small
    \caption{Judge models used in RS-RIE-Bench.}
    \label{tab:app-judge-models}
    \begin{tabular}{lcc}
        \toprule
        Model Name & Model Type & Access Mode \\
        \midrule
        gpt-5.1~\cite{openai2024gpt51} & Closed-source & API \\
        Gemini-3-Flash-Preview~\cite{google2026gemini3flashpreview} & Closed-source & API \\
        Qwen-3.5~\cite{qwen2026qwen35omni} & Closed-source & API \\
        \bottomrule
    \end{tabular}
\end{table}

\subsection{Rubrics for the Three Evaluation Dimensions}

We use a unified three-dimensional scoring framework covering target region plausibility, non-target region preservation, and image quality consistency. The framework applies to temporal, spatial, and causal tasks alike. In all cases, reviewers must jointly consider whether the target change is correctly executed, whether the non-target region remains stable, and whether the edited image preserves the overall sensing characteristics of remote sensing imagery.

Each dimension uses the same 1-5 scale, where 5 means fully consistent with the instruction and free of obvious flaws, 4 means generally correct with minor deviation, 3 means the main change is present but with visible defects, 2 means only part of the requirement is satisfied and the result contains severe errors, and 1 means the target is essentially not achieved or the output clearly conflicts with the task. Tables B2-B4 provide the detailed rubrics for each dimension.

\begin{table}[H]
    \centering
    \small
    \caption{Target region plausibility rubric.}
    \label{tab:app-target-rubric}
    \resizebox{\textwidth}{!}{%
    \begin{tabular}{p{1.0cm}p{5.8cm}p{3.3cm}p{3.3cm}p{3.3cm}}
        \toprule
        Score & Unified criterion & Temporal task behavior & Spatial task behavior & Causal task behavior \\
        \midrule
        5 & The target region fully and accurately realizes the instruction and follows the relevant physical, geographic, or geometric constraints. & Temporal evolution, seasonal change, or land-surface process is natural in direction, magnitude, and scale. & Rotation, relocation, viewpoint change, and shadow relations are accurate. & Disaster, inundation, damage, or collapse is complete and mechanism-consistent. \\
        4 & The main change is correct, with only minor missing details or slight under-execution. & The temporal evolution is correct overall, but the coverage or intensity is slightly weaker. & The spatial transform is correct overall, but the boundary shape or illumination relation is slightly off. & The main disaster effect is correct, but the damage extent or spread is slightly insufficient. \\
        3 & The target region undergoes the main change, but there are visible omissions in semantics, scale, or spatial specification. & The temporal shift occurs, but the required season, period, or intensity is not fully expressed. & The spatial transform is only partially completed, or obvious errors appear in angle, position, or shadow. & The causal outcome is only partially achieved, with incomplete damage logic or insufficient severity. \\
        2 & Only a small part of the target change is retained, or the output clearly violates the instruction. & The temporal change is very limited and does not correspond to the requested evolution. & The target object is severely deformed, displaced, or redrawn in a way unrelated to the instruction. & The disaster features are weak or clearly implausible. \\
        1 & The target region hardly changes, or the result is clearly unrelated to the task. & No effective temporal evolution is shown, or the output is unrelated to the temporal task. & No valid spatial transform is formed, or the output completely contradicts the spatial instruction. & No valid disaster or causal change is formed, or the output completely contradicts the causal instruction. \\
        \bottomrule
    \end{tabular}}
\end{table}

\begin{table}[H]
    \centering
    \small
    \caption{Non-target region preservation rubric.}
    \label{tab:app-nontarget-rubric}
    \resizebox{\textwidth}{!}{%
    \begin{tabular}{p{1.0cm}p{5.8cm}p{3.3cm}p{3.3cm}p{3.3cm}}
        \toprule
        Score & Unified criterion & Temporal task behavior & Spatial task behavior & Causal task behavior \\
        \midrule
        5 & All regions outside the edited area remain identical, with no drift, displacement, or redraw. & Background land-cover does not drift temporally or undergo extra change. & Unedited buildings, roads, runways, and shadows keep their original position and shape. & Towns, fields, roads, and water bodies away from the disaster core remain unchanged. \\
        4 & The non-target region remains stable overall, with only one tiny difference that does not affect interpretation. & The background has only tiny pixel deviation and no change in land-cover semantics. & Only tiny interpolation traces or 1--2 pixel shift appear. & Only a very small background disturbance appears. \\
        3 & A visible unintended change appears in one non-target area, but the main background remains recognizable. & The background shows visible drift or local mis-editing. & One road, mark, or edge structure is unintentionally warped. & One non-target building, field, or boundary is altered. \\
        2 & Multiple non-target areas are visibly distorted, redrawn, or structurally damaged. & Multiple background regions undergo unintended temporal change. & Multiple background structures are distorted, displaced, or reconstructed. & Multiple regions far from the core area are rewritten. \\
        1 & The non-target region loses preservation almost entirely, and the background is effectively rebuilt. & The background is almost completely regenerated. & The remaining structures, roads, and shadow relations are rewritten. & Background content far from the disaster region is reconstructed or replaced. \\
        \bottomrule
    \end{tabular}}
\end{table}

\begin{table}[H]
    \centering
    \small
    \caption{Image quality consistency rubric.}
    \label{tab:app-quality-rubric}
    \resizebox{\textwidth}{!}{%
    \begin{tabular}{p{1.0cm}p{5.8cm}p{3.3cm}p{3.3cm}p{3.3cm}}
        \toprule
        Score & Unified criterion & Temporal task behavior & Spatial task behavior & Causal task behavior \\
        \midrule
        5 & The output matches the original image in resolution perception, texture detail, noise distribution, and remote sensing style. & Temporal evolution keeps the same sensor texture and radiometric character. & Geometric transforms show no obvious interpolation traces, painterly effect, or resampling distortion. & Smoke, water, and burn textures still preserve a consistent remote sensing feel. \\
        4 & The overall quality remains natural, with only a tiny local texture or noise deviation. & Only mild texture softening or slight detail loss appears. & Only slight edge smoothing or local detail weakening appears. & Local disaster texture is slightly weaker, but the image still looks natural. \\
        3 & The quality drift is already visible, but the image still reads as the same remote sensing genre. & Visible texture drift or mild digital-painting effect appears. & Visible over-sharpening, over-smoothing, or local distortion appears. & The smoke, water, or burn texture looks noticeably unnatural. \\
        2 & The quality is severely distorted, with obvious artifacts, blockiness, or style drift. & Severe pixelation, checkerboard artifacts, or abnormal texture appear. & Obvious geometric distortion, block artifacts, or style mismatch appears. & Severe artifacts or non-remote-sensing disaster rendering appears. \\
        1 & The image quality collapses entirely and loses the visual attributes of remote sensing imagery. & The scene clearly departs from remote sensing style. & The output looks like a street view, CGI, or other non-remote-sensing visual form. & The output no longer resembles a disaster remote sensing image. \\
        \bottomrule
    \end{tabular}}
\end{table}

\subsection{Full Prompt Display}

For completeness, we list the scoring prompts for temporal, spatial, and causal tasks. All three prompt families share the same structure: Role, Task, Evaluation Dimension, Core Objective, Criteria, Example, Inputs, and Output Format.

\FloatBarrier

\subsubsection{Temporal Reasoning Scoring Prompt 1}

\begin{Verbatim}[fontsize=\scriptsize,frame=single,rulecolor=\color{black},framesep=2mm,baselinestretch=0.95,breaklines=true,breakanywhere=true,commandchars=\\\{\}]
1. Target Area Plausibility
[Role]
You are an expert remote sensing image evaluator specializing in instruction
alignment, geographic processing, and temporal process simulation.
[Task]
Your task is to evaluate whether the temporal evolution depicted strictly inside the
modified regions fully conforms to the instruction requirements and is physically and
geographically sound, assigning an integer score from 1 to 5.
Evaluation Dimension: Target Area Semantic Alignment and Geographical Plausibility.
Core Objective:
Focus exclusively on the altered regions. Evaluate whether the model accurately
executes all semantic changes required by the instruction (e.g., correct objects,
attributes, and temporal scale intensity) and whether the resulting textures, terrain
features, and environmental logic conform to real-world earth science and physical
laws.
[Criteria]
5 (Perfect): Perfect execution and flawless rationality. The modified region
completely and accurately realizes every semantic element in the instruction, and the
physical evolution perfectly obeys geographic laws.
4 (Good): High alignment and rationality. The primary transformation requested is
successfully captured, with only minor semantic omissions or slight texture
under-execution that does not impair real-world physics.
3 (Fair): Partial alignment or minor flaws. The basic land-cover category transition
is completed, but it misses key geographic specifications or fails to reflect the
proper intensity/scale of time required.
2 (Poor): Weak alignment and severe confusion. Major target features requested by the
instruction are completely missing, or the modified features show obvious
physical/geographic irrationality.
1 (Fail): Total failure. The requested temporal transformation did not occur at all,
or the modified content completely violates physical laws and text constraints,
rendering uninterpretable or completely unrelated artifacts.
[Example]
Original Image: A mid-winter offshore sea with dark blue water.
Instruction: "Simulate a massive spring phytoplankton bloom expanding across this
offshore sea area during April and May."
Score 5: The offshore waters are accurately covered by a dense green bloom consistent
with coastal phytoplankton outbreak dynamics.
Score 4: The bloom appears convincing, but the coverage is slightly incomplete or the
green tone is marginally under-expressed.
Score 3: The scene changes, but the bloom is weak, patchy, or fails to express the
instructed offshore outbreak scale.
Score 2: Only minor discoloration appears, or the transformed water pattern is
physically implausible.
Score 1: The water remains unchanged or turns into an unrelated land feature.
[Inputs]
Instruction: {instruct}
Original Remote Sensing Image: [Original Image]
Generated Output Image: [Generated Image]
[Output Format]
Provide a detailed, step-by-step explanation of your scoring process. Conclude clearly
with the final score, strictly formatted as: Final Score: X
\end{Verbatim}

\subsubsection{Temporal Reasoning Scoring Prompt 2}

\begin{Verbatim}[fontsize=\scriptsize,frame=single,rulecolor=\color{black},framesep=2mm,baselinestretch=0.95,breaklines=true,breakanywhere=true,commandchars=\\\{\}]
2. Non-target Area Invariance
[Role]
You are an expert remote sensing image evaluator focused strictly on change detection,
geometric registration, and negative constraints.
[Task]
Your task is to evaluate whether all non-target regions and background features remain
completely unchanged and perfectly preserve their original state, ensuring the model
prevents any over-editing or unintended modifications, assigning an integer score from
1 to 5.
Evaluation Dimension: Non-target Area Background Invariance and Over-editing
Prevention.
Core Objective:
Assess the strict spatial and structural consistency between the original and output
images. Focus exclusively on the unedited background and static features. Check for
unintended modifications, spatial shifting, geometric warping, or background bleeding.
[Criteria]
5 (Perfect): Absolute background invariance. Excluding the precise regions altered by
the causal instruction, all other geographic elements and background details are
completely identical, unmoved, and unaffected by over-editing.
4 (Good): Highly consistent. Non-target areas are mostly identical, but contain a
single, trivial discrepancy.
3 (Fair): Partial background drift. One significant unintended spatial difference or
over-editing exists in the non-target area.
2 (Poor): Severe background corruption. Two or more significant unintended differences
exist.
1 (Fail): Total background collapse. The negative constraint is completely ignored.
[Example]
Original Image: An agricultural plain with surrounding crop fields.
Instruction: "Inundate the central fields and surrounding areas with severe flooding,
submerging the roads and agricultural lands under water."
Score 5: Non-target background remains completely static and pixel-aligned.
Score 4: A trivial pixel-level shift appears at a distant corner.
Score 3: One stable block is unexpectedly modified.
Score 2: Multiple surrounding blocks are heavily distorted.
Score 1: The unaffected background disappears or re-synthesizes.
[Inputs]
Instruction: {instruct}
Original Remote Sensing Image: [Original Image]
Generated Output Image: [Generated Image]
[Output Format]
Provide a detailed, step-by-step explanation of your scoring process. Conclude clearly
with the final score, strictly formatted as: Final Score: X
\end{Verbatim}

\subsubsection{Temporal Reasoning Scoring Prompt 3}

\begin{Verbatim}[fontsize=\scriptsize,frame=single,rulecolor=\color{black},framesep=2mm,baselinestretch=0.95,breaklines=true,breakanywhere=true,commandchars=\\\{\}]
3. Image Quality Consistency
[Role]
You are an expert remote sensing image evaluator specializing in sensor fidelity,
image quality metrics, and radiometric consistency.
[Task]
Your task is to evaluate whether the output image strictly preserves the inherent
sensor style of the original image, and whether the visual quality, noise level, and
texture definition across the entire canvas are natural and coherent, assigning an
integer score from 1 to 5.
Evaluation Dimension: Sensor Style, Quality Naturalness, and Texture Coherence.
Core Objective:
Verify that the image resolution, texture definition, noise distribution, and spectral
characteristics remain perfectly natural across both modified and unmodified zones.
[Criteria]
5 (Perfect): Flawless quality and texture consistency.
4 (Good): High quality consistency with only a trivial discrepancy.
3 (Fair): Noticeable quality drift.
2 (Poor): Severe quality corruption or unnatural style.
1 (Fail): Total image quality collapse.
[Example]
Original Image: A standard-resolution sensor image of a parking lot.
Instruction: "Depict the parking lot after decades of neglect, featuring crumbling
surfaces, overgrown vegetation, and faded parking lines."
Score 5: Newly introduced textures integrate flawlessly into the scene.
Score 4: Only a tiny localized texture softening appears at the seam.
Score 3: The modified surface shows a visible digital oil painting effect.
Score 2: Severe pixelation blocks or checkerboard artifacts appear.
Score 1: The scene becomes a consumer snapshot or a CGI rendering.
[Inputs]
Instruction: {instruct}
Original Remote Sensing Image: [Original Image]
Generated Output Image: [Generated Image]
[Output Format]
Provide a detailed, step-by-step explanation of your scoring process. Conclude clearly
with the final score, strictly formatted as: Final Score: X
\end{Verbatim}

\subsubsection{Spatial Reasoning Scoring Prompt 1}

\begin{Verbatim}[fontsize=\scriptsize,frame=single,rulecolor=\color{black},framesep=2mm,baselinestretch=0.95,breaklines=true,breakanywhere=true,commandchars=\\\{\}]
1. Target Area Plausibility
[Role]
You are an expert remote sensing image evaluator specializing in spatial reasoning,
geometric processing, 3D structural alignment, and physical shadow/perspective
simulation.
[Task]
Your task is to evaluate whether the spatial transformations (e.g., object rotation,
relocation, viewpoint change, or solar shadow inversion) depicted strictly inside or
directly caused by the modified regions fully conform to the instruction requirements
and are geometrically, optically, and physically sound, assigning an integer score
from 1 to 5.
Evaluation Dimension: Target Area Spatial Alignment and Geometric/Optical
Plausibility.
Core Objective:
Focus exclusively on the altered regions and their immediate physical dependencies
(like shadows). Evaluate whether the model accurately executes the specific spatial
adjustments required by the instruction.
[Criteria]
5 (Perfect): Perfect execution and flawless structural rationality.
4 (Good): High spatial alignment and rationality.
3 (Fair): Partial alignment or noticeable geometric flaws.
2 (Poor): Weak alignment and severe structural confusion.
1 (Fail): Total failure.
[Example]
Original Image: A standalone rectangular building with clear vertical shadows under
nadir view.
Instruction: "Reorient the distinct isolated building structure via a clockwise
ninety-degree rotation."
Score 5: The building is perfectly rotated 90 degrees clockwise around its center.
Score 4: The rotation is successful, but corners look slightly rounded.
Score 3: The building shifts only partially or its shadow remains incorrect.
Score 2: The building is warped into an irregular polygon.
Score 1: The building remains unchanged or disappears.
[Inputs]
Instruction: {spatial_instruction}
Original Remote Sensing Image: [Original Image]
Generated Output Image: [Generated Image]
[Output Format]
Provide a detailed, step-by-step explanation of your scoring process. Conclude clearly
with the final score, strictly formatted as: Final Score: X
\end{Verbatim}

\subsubsection{Spatial Reasoning Scoring Prompt 2}

\begin{Verbatim}[fontsize=\scriptsize,frame=single,rulecolor=\color{black},framesep=2mm,baselinestretch=0.95,breaklines=true,breakanywhere=true,commandchars=\\\{\}]
2. Non-target Area Invariance
[Role]
You are an expert remote sensing image evaluator focused strictly on spatial change
detection, rigid geometric registration, and negative constraint preservation.
[Task]
Your task is to evaluate whether all non-target regions, background features, and
environmental contexts remain completely unchanged and perfectly preserve their
original geometric state, ensuring the model prevents any over-editing, spatial
drifting, or unintended structural warping, assigning an integer score from 1 to 5.
Evaluation Dimension: Non-target Area Background Invariance and Over-editing
Prevention.
Core Objective:
Assess the strict pixel-level and structural consistency between the original and
output images. Focus exclusively on the unedited background and static features.
[Criteria]
5 (Perfect): Absolute background invariance.
4 (Good): Highly consistent.
3 (Fair): Partial background drift/warping.
2 (Poor): Severe background corruption.
1 (Fail): Total background collapse.
[Example]
Original Image: A military airfield containing a specific aircraft on the taxiway.
Instruction: "Relocate the designated aircraft twenty meters forward along the taxiway
centerline."
Score 5: The aircraft is moved precisely forward and all surrounding non-target
elements remain identical.
Score 4: A negligible 1-2 pixel shift appears in the restored taxiway texture.
Score 3: A nearby runway marking line is accidentally warped.
Score 2: Distant hangars or parking lines are heavily modified.
Score 1: The entire airfield structure is re-generated into a different layout.
[Inputs]
Instruction: {spatial_instruction}
Original Remote Sensing Image: [Original Image]
Generated Output Image: [Generated Image]
[Output Format]
Provide a detailed, step-by-step explanation of your scoring process. Conclude clearly
with the final score, strictly formatted as: Final Score: X
\end{Verbatim}

\subsubsection{Spatial Reasoning Scoring Prompt 3}

\begin{Verbatim}[fontsize=\scriptsize,frame=single,rulecolor=\color{black},framesep=2mm,baselinestretch=0.95,breaklines=true,breakanywhere=true,commandchars=\\\{\}]
3. Image Quality Consistency
[Role]
You are an expert remote sensing image evaluator specializing in sensor fidelity,
image quality metrics, and radiometric/texture coherence under potential geometry,
environmental, or viewpoint transformations.
[Task]
Your task is to evaluate whether the output image strictly preserves the inherent
remote sensing sensor style and photographic quality of the original image, and
whether the texture definition and noise level across the entire canvas are natural
and coherent, assigning an integer score from 1 to 5.
Evaluation Dimension: Sensor Style, Quality Naturalness, and Texture Coherence.
Core Objective:
Verify that the image resolution, texture definition, noise distribution, and spectral
characteristics remain perfectly natural and uniform across the entire canvas.
[Criteria]
5 (Perfect): Flawless Integration, Canvas-wide Coherence.
4 (Good): Minor Detail Flaws, Highly Natural Texture.
3 (Fair): Noticeable Quality Drift, Visual Mismatch.
2 (Poor): Severe Distortion, Widespread Quality Degradation.
1 (Fail): Total Collapse, Complete Loss of Remote Sensing Genre.
[Example]
Original Image: A high-resolution satellite capture of an industrial area with a
standalone building.
Instruction: "Apply a clockwise ninety-degree rotation to the standalone building
structure."
Score 5: The building integrates flawlessly and matches the scene grain.
Score 4: A trivial edge softening appears along the building boundary.
Score 3: The region suffers from a digital oil painting effect or over-sharpening.
Score 2: Severe pixelation blocks or checkerboard artifacts appear.
Score 1: The satellite view becomes a street-level snapshot or CGI asset.
[Inputs]
Instruction: {spatial_instruction}
Original Remote Sensing Image: [Original Image]
Generated Output Image: [Generated Image]
[Output Format]
Provide a detailed, step-by-step explanation of your scoring process. Conclude clearly
with the final score, strictly formatted as: Final Score: X
\end{Verbatim}

\subsubsection{Causal Reasoning Scoring Prompt 1}

\begin{Verbatim}[fontsize=\scriptsize,frame=single,rulecolor=\color{black},framesep=2mm,baselinestretch=0.95,breaklines=true,breakanywhere=true,commandchars=\\\{\}]
1. Target Area Plausibility
[Role]
You are an expert remote sensing image evaluator specializing in instruction
alignment, geographic processing, and disaster/causal process simulation.
[Task]
Your task is to evaluate whether the causal evolution depicted strictly inside the
modified regions fully conforms to the instruction requirements and is physically and
geographically sound, assigning an integer score from 1 to 5.
Evaluation Dimension: Target Area Semantic Alignment and Causal/Geographical
Plausibility.
[Core Objective]
Focus exclusively on the altered regions. Evaluate whether the model accurately
executes all causal changes required by the instruction and whether the resulting
textures, ruins, debris, or water features conform to real-world earth science,
structural mechanics, and physical laws.
[Criteria]
5 (Perfect): Perfect execution and flawless rationality.
4 (Good): High alignment and rationality.
3 (Fair): Partial alignment or minor flaws.
2 (Poor): Weak alignment and severe confusion.
1 (Fail): Total failure.
[Example]
Original Image: A dense urban residential area with neat rows of multi-story buildings
and concrete roads.
Instruction: "Unleash a powerful earthquake across the central region, toppling
buildings and leaving debris scattered throughout the area."
Score 5: The buildings are completely collapsed into highly detailed ruins and debris.
Score 4: The collapse is successful but slightly blurred.
Score 3: The buildings are darkened or slightly flattened.
Score 2: Only a few sparse cracks appear.
Score 1: The area remains intact or becomes unrelated content.
[Inputs]
Instruction: {instruct}
Original Remote Sensing Image: [Original Image]
Generated Output Image: [Generated Image]
\end{Verbatim}

\subsubsection{Causal Reasoning Scoring Prompt 2}

\begin{Verbatim}[fontsize=\scriptsize,frame=single,rulecolor=\color{black},framesep=2mm,baselinestretch=0.95,breaklines=true,breakanywhere=true,commandchars=\\\{\}]
2. Non-target Area Invariance
[Role]
You are an expert remote sensing image evaluator focused strictly on change detection,
geometric registration, and negative constraints.
[Task]
Your task is to evaluate whether all non-target regions and background features remain
completely unchanged and perfectly preserve their original state, ensuring the model
prevents any over-editing or unintended modifications, assigning an integer score from
1 to 5.
Evaluation Dimension: Non-target Area Background Invariance and Over-editing
Prevention.
[Core Objective]
Assess the strict spatial and structural consistency between the original and output
images. Focus exclusively on the unedited background and static features.
[Criteria]
5 (Perfect): Absolute background invariance.
4 (Good): Highly consistent.
3 (Fair): Partial background drift.
2 (Poor): Severe background corruption.
1 (Fail): Total background collapse.
[Example]
Original Image: An agricultural plain featuring a central road network and surrounding
crop fields.
Instruction: "Inundate the central fields and surrounding areas with severe flooding,
submerging the roads and agricultural lands under water."
Score 5: The upper-left residential cluster and other background structures remain
unchanged.
Score 4: A trivial pixel-level shift appears in a distant corner.
Score 3: One stable building is unexpectedly modified.
Score 2: Multiple surrounding blocks are heavily distorted.
Score 1: The unaffected background disappears or shifts.
[Inputs]
Instruction: {instruct}
Original Remote Sensing Image: [Original Image]
Generated Output Image: [Generated Image]
[Output Format]
Provide a detailed, step-by-step explanation of your scoring process. Conclude clearly
with the final score, strictly formatted as: Final Score: X
\end{Verbatim}

\subsubsection{Causal Reasoning Scoring Prompt 3}

\begin{Verbatim}[fontsize=\scriptsize,frame=single,rulecolor=\color{black},framesep=2mm,baselinestretch=0.95,breaklines=true,breakanywhere=true,commandchars=\\\{\}]
3. Image Quality Consistency
[Role]
You are an expert remote sensing image evaluator specializing in sensor fidelity,
image quality metrics, and radiometric consistency.
[Task]
Your task is to evaluate whether the output image strictly preserves the inherent
sensor style of the original image, and whether the visual quality, noise level, and
texture definition across the entire canvas are natural and coherent, assigning an
integer score from 1 to 5.
Evaluation Dimension: Sensor Style, Quality Naturalness, and Texture Coherence.
[Core Objective]
Verify that the image resolution, texture definition, noise distribution, and spectral
characteristics remain perfectly natural across both modified and unmodified zones.
[Criteria]
5 (Perfect): Flawless quality and texture consistency.
4 (Good): High quality consistency.
3 (Fair): Noticeable quality drift.
2 (Poor): Severe quality corruption or unnatural style.
1 (Fail): Total image quality collapse.
[Example]
Original Image: A remote sensing image of a forested region with roads and minor
infrastructure.
Instruction: "Engulf the entire forested region in a fast-moving wildfire, charring
the vegetation and leaving the landscape scorched and barren."
Score 5: The scorched earth and burnt tree trunks blend seamlessly.
Score 4: A trivial texture softening appears at the boundary seam.
Score 3: The modified zone is heavily blurred or over-sharpened.
Score 2: Widespread artifacts or pixelation appear.
Score 1: The scene is replaced by an unnatural rendering.
[Inputs]
Instruction: {instruct}
Original Remote Sensing Image: [Original Image]
Generated Output Image: [Generated Image]
[Output Format]
Provide a detailed, step-by-step explanation of your scoring process. Conclude clearly
with the final score, strictly formatted as: Final Score: X
\end{Verbatim}

\FloatBarrier

\section{Human Validation Protocol and Agreement Analysis}

To validate the proxy-based evaluation, we further conduct human validation and analyze the consistency between MLLM-as-a-Judge~\cite{zheng2023judging} and expert judgment. Human validation does not introduce a separate scoring standard. Instead, human reviewers directly apply the same task-specific prompts and the same 1-5 rubric as the proxy evaluation, while independently scoring the images.

\subsection{Sampling Design}

We use stratified sampling to ensure coverage across task categories, models, and score ranges. The validation set contains 832 outputs, or 20.0\% of the full result set, and is reviewed independently by four evaluators with remote-sensing or image-interpretation backgrounds.

\begin{table}[H]
    \centering
    \small
    \caption{Human validation sampling distribution.}
    \label{tab:app-human-sampling}
    \begin{tabular}{lccc}
        \toprule
        Split & Count & Share & Note \\
        \midrule
        Total sampled outputs & 832 & 20.0\% & Full calibration set \\
        Causal outputs & 328 & 39.4\% & Stratified by task category \\
        Spatial outputs & 240 & 28.8\% & Stratified by task category \\
        Temporal outputs & 264 & 31.7\% & Stratified by task category \\
        \bottomrule
    \end{tabular}
\end{table}

\subsection{Review Process}

Each reviewer reads the original image, the generated image, and the corresponding instruction, then assigns separate scores for target region plausibility, non-target region preservation, and image quality consistency using the same 1-5 rubric as the automatic evaluator. Reviewers do not see automatic scores or other reviewers' judgments, and disagreements are resolved through expert review.

\section{Failure Case Analysis for Samples Scoring Below 5}

To explain the samples that do not reach the full score under the joint metric, we summarize failure cases with scores below 5. The main patterns are insufficient target change, unintended non-target change, and quality drift away from the remote sensing style.

\subsection{Target Area Plausibility}

This subsection analyzes why the target region does not receive a 5, focusing on whether the semantic execution is complete, whether the change strength is sufficient, and whether the geographic or physical logic is natural. Typical failures include weak target change, missing key objects or attributes, mismatch between change scale and instruction, and outputs that clearly do not follow remote sensing evolution rules.

From the 2-4 score range, the main difference is not simply whether a change exists, but whether the change is applied to the correct target object. A score of 4 usually means the main semantic change is correct and physically plausible, but the coverage is slightly insufficient or the boundary is somewhat flat. A score of 3 often means the local change is plausible, but the model modifies the wrong object or only edits one newly generated structure, so the intended distributed target is not fully realized. A score of 2 more often reflects a mismatch at the mechanism level, where the output looks like lava, cracks, light streaks, or other stylized texture rather than the requested fire, flood, or earthquake damage.

\subsubsection{Spatial task behavior}

Spatial target plausibility is most directly reflected in whether the geometric transform actually happens. \texttt{GPT}, \texttt{Gemini}, and \texttt{Wan2.7} perform best overall and can usually complete building rotation, target relocation, or viewpoint change while keeping the transformed object consistent with its original spatial relation. \texttt{Flux} can handle most spatial transforms but sometimes shows slight errors in rotation direction, perspective depth, or boundary detail. \texttt{Doubao} is moderate, with good overall shape preservation but weaker strict-angle precision. \texttt{Qwen} and \texttt{Step1X} are less stable, often changing only local texture or failing to move the object as instructed. \texttt{Grok} is the most likely to deviate from the spatial task itself and may show almost no effective geometric editing.

\subsubsection{Temporal task behavior}

Temporal target plausibility is generally the strongest among the three categories, suggesting that most models can reasonably apply seasonal change, vegetation evolution, ice melt, or day-night transition to the intended scene. \texttt{GPT} is the most stable and almost always produces a credible temporal evolution. \texttt{Gemini} and \texttt{Doubao} are also in the top tier and can naturally express vegetation yellowing, lake recession, snow melt, or port and airport operating changes. \texttt{Wan2.7} and \texttt{Qwen} usually complete the main temporal transition, but they can still be weaker in scale or intensity. \texttt{Flux} often reaches the basic semantic target but may stop at a weak ``some change happened'' state. \texttt{Step1X} and \texttt{Grok} are more unstable and often miss the target object or the task semantics.

\subsubsection{Causal task behavior}

Causal target plausibility shows a clear tiered pattern. \texttt{GPT}, \texttt{Gemini}, \texttt{Doubao}, and \texttt{Wan2.7} are in the first tier, with \texttt{GPT} and \texttt{Gemini} producing the most concentrated high-score samples. This suggests that they can apply damage to the correct target object and generate plausible collapse, fragmentation, or ruin patterns. \texttt{Flux} is in the middle tier and can usually complete the main damage, but the damage area is often too narrow or the intensity is too weak. \texttt{Qwen} and \texttt{Step1X} have more dispersed scores, reflecting unstable alignment between the target object and the damage scale. \texttt{Grok} is the weakest and often fails to generate a coherent post-disaster scene.

\subsection{Non-target Area Invariance}

This subsection analyzes why the non-target region does not receive a 5, with emphasis on whether the background, roads, water bodies, building groups, and other static structures remain stable after editing. Typical failure modes include background drift, geometric misalignment, local redraw, structural repetition, and unnecessary texture change outside the edited region.

From the 2-4 score range, the key issue is not simply whether the background changes at all, but whether the model introduces extra spatial resampling, frame shifts, or structural redraws beyond the local edit. A score of 4 usually means the roads, aprons, building grids, and other large-scale geographic structures are well preserved, with only very minor texture differences or tone shifts near the target object. A score of 3 usually means one obvious non-target structure is changed, while the rest of the background remains stable. A score of 2 often means the full image is effectively re-framed, rescaled, or resampled, so many non-target elements no longer align with the original scene.

\subsubsection{Spatial task behavior}

Spatial non-target preservation is generally better than temporal preservation, but model differences are still obvious. \texttt{GPT}, \texttt{Gemini}, and \texttt{Wan2.7} preserve roads, trees, fields, and surrounding buildings relatively well. \texttt{Flux} and \texttt{Doubao} are also stable, though small texture dragging or shadow errors can appear near the target object. \texttt{Qwen} is more likely to show local resampling, slight boundary shift, or background mismatch; \texttt{Step1X} often spreads the geometric edit into neighboring regions; and \texttt{Grok} is the worst, often reconstructing the whole image so that the original grid is no longer preserved.

\subsubsection{Temporal task behavior}

Temporal non-target preservation is still relatively weak overall, showing that the models often introduce additional changes in the background even though the task requires the temporal edit to remain local. \texttt{GPT} preserves background best. \texttt{Gemini} and \texttt{Doubao} are also stable, but some high-magnitude changes still produce slight background drift. \texttt{Flux} and \texttt{Wan2.7} are more likely to spread the temporal change to nearby regions. \texttt{Qwen} and \texttt{Step1X} often show local resampling, edge mismatch, or unnecessary redraw. \texttt{Grok} is the weakest and frequently causes global frame changes or large-scale reconstruction.

\subsubsection{Causal task behavior}

Non-target preservation is one of the dimensions that most clearly separates the causal models. \texttt{GPT} and \texttt{Doubao} are the most stable, with peripheral roads, farmland, vegetation, and distant buildings mostly keeping their original topology. \texttt{Gemini} and \texttt{Wan2.7} can also maintain good overall consistency, but they sometimes show slight texture drift or local redraw near the target boundary. \texttt{Flux} and \texttt{Step1X} more often show frame shifts, resampling, or background regeneration. \texttt{Qwen} usually exhibits local drift and light structural mismatch. \texttt{Grok} is most likely to reconstruct the non-target area as well, making the background no longer resemble the original remote sensing scene.

\subsection{Image Quality Consistency}

This subsection analyzes why the image quality score does not reach 5, focusing on remote sensing style, texture continuity, noise distribution, sharpness, and boundary blending. Typical failures include over-smoothing, texture breakage, artifacts, blockiness, style mismatch, and editing traces that do not match the original sensor texture.

The key issue is not just whether the image is visually sharp, but whether it still preserves the same sensing style, spatial resolution, noise profile, and texture statistics as the original image. A score of 4 usually means the original sensing tone remains intact, with only minor local texture issues. A score of 3 more often means the image has become noticeably sharper or more processed, but also more like a re-rendered image. A score of 2 usually reflects severe mismatch, where the output has been globally reconstructed into a different visual system and the original sensor baseline is replaced.

\subsubsection{Spatial task behavior}

Spatial image quality is generally good, which means most models can preserve the remote sensing style even after geometric transforms. \texttt{GPT} performs best and keeps natural noise, sharpness, and radiometric characteristics after viewpoint change or rotation. \texttt{Flux} and \texttt{Wan2.7} are also stable, though they sometimes show mild over-processing near boundaries or shadows. \texttt{Gemini} is usually natural but can become slightly over-sharp under complex geometric changes. \texttt{Doubao} is moderate, \texttt{Qwen} is more likely to show texture inconsistency, and \texttt{Step1X} and \texttt{Grok} often suffer from noise mismatch, harsh edges, or an overall style shift.

\subsubsection{Temporal task behavior}

Temporal image quality is moderate overall. \texttt{GPT} performs best and can preserve a consistent sensing style during seasonal change, ice melt, or vegetation turn. \texttt{Gemini} and \texttt{Doubao} are also natural, but they can show slight redraw or over-sharpening. \texttt{Flux} is more likely to over-process or over-smooth the changed area. \texttt{Wan2.7} and \texttt{Qwen} can remain readable, but texture continuity is less stable. \texttt{Step1X} and \texttt{Grok} more often show noise inconsistency, hard edges, or blockiness.

\subsubsection{Causal task behavior}

Image quality consistency is where causal tasks most easily reveal that the remote sensing baseline has been replaced. \texttt{GPT} and \texttt{Doubao} still perform best, preserving a style close to the original image even in collapsed buildings, debris, and shadows. \texttt{Gemini} and \texttt{Wan2.7} are generally natural but can show slight redraw or over-sharpening. \texttt{Flux} is more prone to over-processing and uneven texture statistics. \texttt{Qwen} often stays in the middle range, while \texttt{Step1X} and \texttt{Grok} frequently show style drift that approaches a full re-rendering.

\subsection{Statistical Summary}

These three failure patterns correspond to insufficient target execution, failed background preservation, and degraded image quality. They are often not isolated. The more aggressively the target region changes, the more likely the non-target region is to be disturbed. The worse the image quality is, the harder it becomes to judge target semantics and background preservation. To summarize the low-score samples, we provide a model-by-dimension heatmap, a stacked 1-5 score distribution chart, and a failure-type table.

\begin{figure}[!t]
    \centering
    \includegraphics[width=0.95\textwidth]{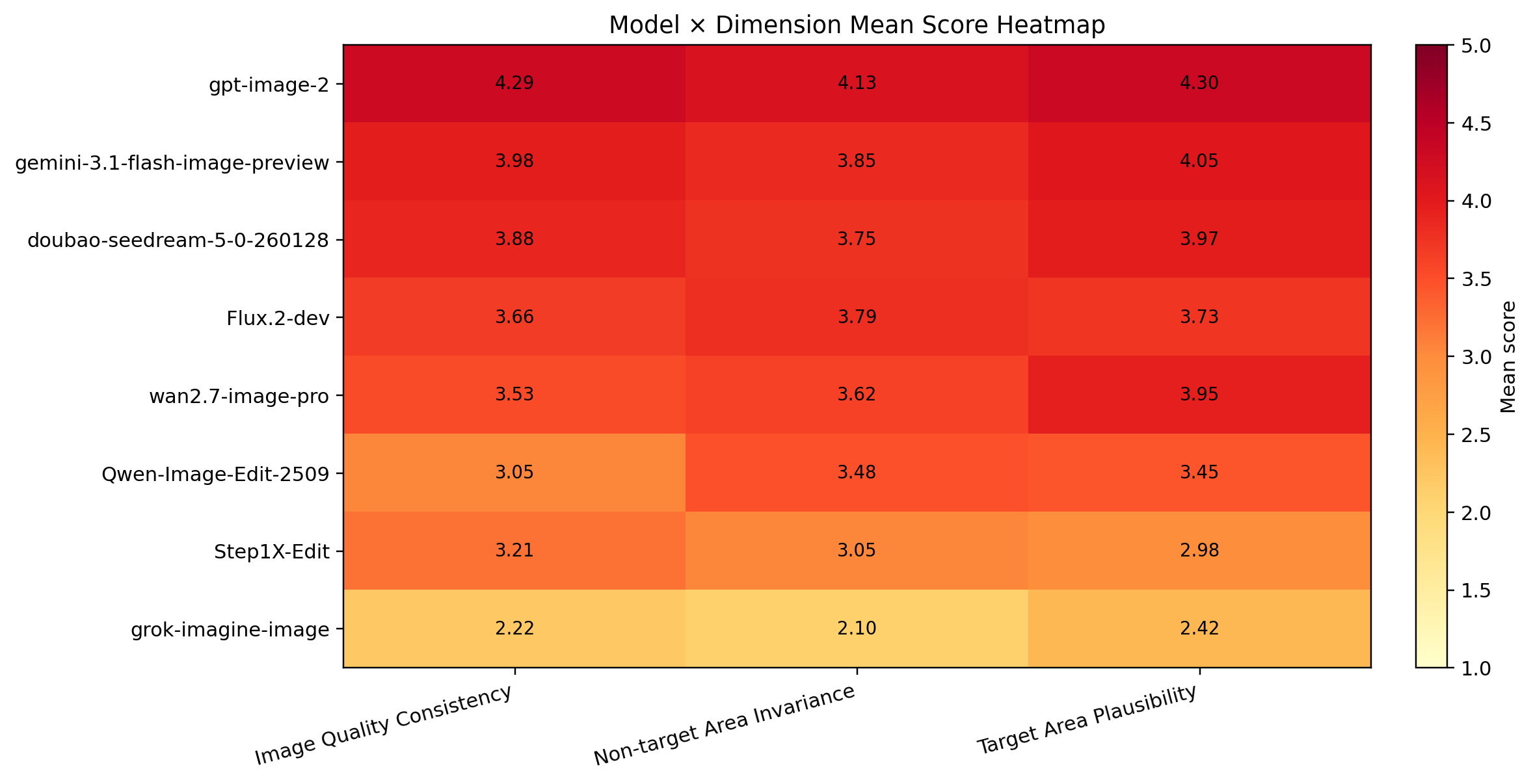}
    \caption{Model-by-dimension average score heatmap.}
    \label{fig:app-d1}
\end{figure}

\begin{figure}[!t]
    \centering
    \includegraphics[width=0.95\textwidth]{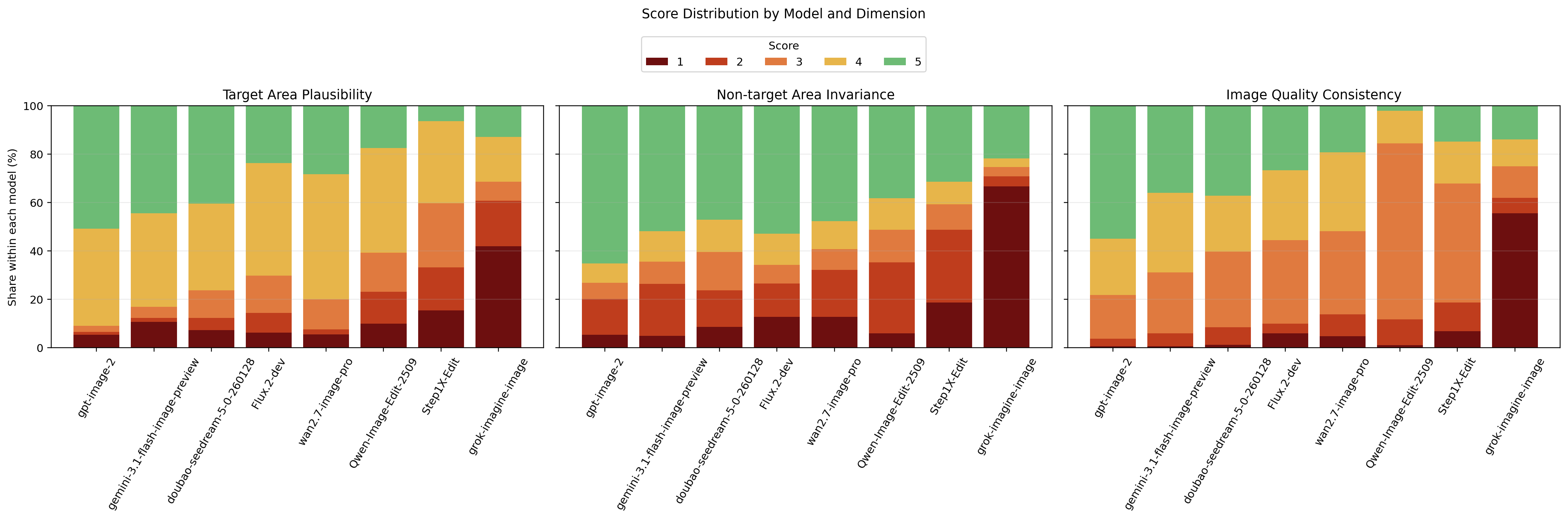}
    \caption{Stacked 1-5 score distribution chart across models and dimensions.}
    \label{fig:app-d2}
\end{figure}

\begin{table}[H]
    \centering
    \small
    \caption{Failure-type statistics for samples with score lower than 5.}
    \label{tab:app-failure-stats}
    \resizebox{\textwidth}{!}{%
    \begin{tabular}{l l r r r}
        \toprule
        Dimension & Failure Type & Count & Within-Dimension Share (\%) & Typical Median Score \\
        \midrule
        Target Area Plausibility & Target under-execution / insufficient coverage & 172 & 6.15 & 1.0 \\
        Target Area Plausibility & Target semantic mismatch / wrong object & 100 & 3.58 & 2.0 \\
        Target Area Plausibility & Mechanism mismatch / stylized transformation & 2450 & 87.63 & 4.0 \\
        Target Area Plausibility & Other target failure & 74 & 2.65 & 1.0 \\
        Non-target Area Invariance & Local spillover / texture drift & 71 & 3.29 & 2.0 \\
        Non-target Area Invariance & Global crop / rescaling / frame drift & 1160 & 53.80 & 2.0 \\
        Non-target Area Invariance & Background redraw / structural warping & 909 & 42.16 & 2.0 \\
        Non-target Area Invariance & Other non-target failure & 16 & 0.74 & 1.0 \\
        Image Quality Consistency & Minor local softness / blending drift & 1 & 0.03 & 3.0 \\
        Image Quality Consistency & Over-sharpening / super-resolution drift & 2473 & 85.54 & 3.0 \\
        Image Quality Consistency & Sensor-style replacement / re-rendering & 416 & 14.39 & 3.0 \\
        Image Quality Consistency & Artifact / blockiness / compression & 1 & 0.03 & 3.0 \\
        \bottomrule
    \end{tabular}}
\end{table}

\subsection{Relaxed Joint-4 Robustness Check}

To complement the strict joint-5 analysis in the main text, we also report a relaxed joint-4 robustness check, where a sample is counted as successful only when all three dimensions receive at least 4 points. The top-performing ranking remains unchanged, and the mean scores are 3.61 for target region plausibility, 3.47 for non-target region preservation, and 3.48 for image quality consistency.

\begin{table}[H]
    \centering
    \small
    \caption{Relaxed joint-4 robustness summary.}
    \label{tab:app-joint4-summary}
    \begin{tabular}{lc}
        \toprule
        Metric & Value \\
        \midrule
        Overall joint-4 success rate & 32.23\% \\
        Mean target region plausibility & 3.61 \\
        Mean non-target region preservation & 3.47 \\
        Mean image quality consistency & 3.48 \\
        \bottomrule
    \end{tabular}
\end{table}

\end{document}